\documentclass[pdflatex,sn-nature]{sn-jnl}
\usepackage{xr-hyper}
\usepackage{graphicx}%
\usepackage{multirow}%
\usepackage{amsmath,amssymb,amsfonts}%
\usepackage{amsthm}%
\usepackage{mathrsfs}%
\usepackage[title]{appendix}%
\usepackage{xcolor}%
\usepackage{textcomp}%
\usepackage{manyfoot}%
\usepackage{booktabs}%
\usepackage{algorithm}%
\usepackage{algorithmicx}%
\usepackage{algpseudocode}%
\usepackage{listings}%
\usepackage{titlesec}
\titleformat{\paragraph}[runin]{\normalfont\bfseries\itshape}{}{0em}{}[]
\theoremstyle{thmstyleone}%
\theoremstyle{thmstyletwo}%

\theoremstyle{thmstylethree}%

\begin{document}

\title[Large-Scale Slice-wise Self-Supervised Brain MRI Foundation Model]
{Frozen Brain MRI Foundation Model Representations Generalize Across Clinical Tasks}


\author[1,3]{\fnm{Yizhou} \sur{Wu}}

\author[2]{\fnm{Shansong} \sur{Wang}}

\author[1,4]{\fnm{Yuheng} \sur{Li}}
\author[2]{\fnm{Mojtaba}  \sur{Safari}}
\author[1,5]{\fnm{Mingzhe} \sur{Hu}}
\author[1]{\fnm{Chih-Wei}  \sur{Chang}}
\author[6]{\fnm{Harini}  \sur{Veeraraghavan}}
\author*[2]{\fnm{Xiaofeng} \sur{Yang}}\email{Xiaofeng.Yang@bsd.uchicago.edu}


\affil[1]{\orgdiv{Department of Radiation Oncology and Winship Cancer Institute},
\orgname{Emory University},
\orgaddress{\city{Atlanta}, \state{GA}, \country{USA}}}

\affil[2]{\orgdiv{Department of Radiation and Cellular Oncology},
\orgname{The University of Chicago},
\orgaddress{\city{Chicago}, \state{IL}, \country{USA}}}

\affil[3]{\orgdiv{Department of Electrical and Computer Engineering},
\orgname{Georgia Institute of Technology},
\orgaddress{\city{Atlanta}, \state{GA}, \country{USA}}}

\affil[4]{\orgdiv{Department of Biomedical Engineering},
\orgname{Georgia Institute of Technology},
\orgaddress{\city{Atlanta}, \state{GA}, \country{USA}}}

\affil[5]{\orgdiv{Department of Biomedical Informatics},
\orgname{Emory University},
\orgaddress{\city{Atlanta}, \state{GA}, \country{USA}}}

\affil[6]{\orgdiv{Department of Medical Physics},
\orgname{Memorial Sloan Kettering Cancer Center},
\orgaddress{\city{New York}, \state{NY}, \country{USA}}}


\abstract{
Deep learning for brain MRI in clinical settings is constrained by the cost of 
expert annotation and the heterogeneity of acquisition across institutions. We 
present BrainDINO, a self-supervised foundation model pretrained on approximately 
6.6 million unlabeled axial slices from 20 datasets spanning diverse patient 
populations, disease entities, scanner vendors, and acquisition protocols. Using 
a frozen encoder with lightweight task heads that update as little as 0.6\% of model
parameters for non-dense tasks (and up to 22.6\% for dense segmentation), BrainDINO supports tumor segmentation, neurodegenerative and
neurodevelopmental disease classification, brain age estimation, post-stroke 
temporal prediction, molecular status prediction, MRI sequence classification, 
and survival risk stratification. Across all tasks, BrainDINO matches or exceeds 
natural-image and MRI-specific self-supervised baselines, with the largest gains 
emerging under low-label regimes representative of real-world deployment. The 
learned representations are anatomically organized, pathology-sensitive, and 
stable under acquisition-related perturbations, establishing a scalable, 
parameter-efficient foundation for data-efficient brain MRI analysis. Code is available at https://github.com/mclwu22/BrainDINO
}




\maketitle

\section{Introduction}

Deep learning has transformed brain magnetic resonance imaging (MRI) across diverse clinical applications, including tumor segmentation\cite{menze2014multimodal,bakas2017advancing,isensee2018nnu}, Neurodevelopmental and Neurodegenerative Classification~\cite{suk2013deep,ebrahimi2021convolutional,dardouri2025efficient}, biomarker estimation such as brain age prediction~\cite{cole2017predicting,peng2021accurate,bashyam2020mri}, and prognostic modeling in neuro-oncology~\cite{kickingereder2016radiomic,macyszyn2015imaging,weninger2018segmentation}. Nevertheless, most existing approaches remain task-specific, requiring substantial labeled data that are often limited and costly to obtain~\cite{isensee2021nnu, menze2014multimodal, liu2023deep}. As a result, learned representations are fragmented, limiting reuse, reducing data efficiency under limited annotation, and restricting domain generalization across heterogeneous clinical settings~\cite{guan2021domain,yoon2024domain,eidex2024deep}.

Similar challenges in natural image domains have been addressed by self-supervised learning (SSL), which learns transferable representations from large-scale unlabeled data. SSL has evolved from contrastive learning, which emphasizes instance-level discrimination~\cite{chen2020simple,chen2020improved,grill2020bootstrap}, to masked reconstruction approaches that capture fine-grained structures~\cite{he2022masked,xie2022simmim,bao2021beit}, and more recently to self-distillation frameworks such as DINO~\cite{caron2021emerging} and DINOv2~\cite{oquab2023dinov2} that align representations across views without explicit supervision. A comparable evolution has been observed in brain MRI, including contrastive frameworks such as 3D SimCLR Foundation~\cite{kaczmarek2025building} and BrainIAC~\cite{tak2026generalizable}, reconstruction-based approaches including AMAES~\cite{munk2024amaes}, BM-MAE~\cite{robinet2025multimodal}, and BrainMVP~\cite{rui2025multi}, self-distillation adaptation such as BrainFound~\cite{mazher2025towards}, and emerging generative or hybrid paradigms such as GenBrain~\cite{yang2025genbrain}.

Despite these advances, existing brain MRI SSL paradigms capture only partial aspects of the representation space. Contrastive approaches primarily encode global semantics, while reconstruction-based methods preserve fine-grained local structures. More critically, all current brain MRI SSL approaches, including contrastive, reconstruction-based, and self-distillation-based methods, typically require full-network fine-tuning to achieve strong downstream performance~\cite{tak2026generalizable, robinet2025multimodal, rui2025multi, mazher2025towards}. This reliance on heavy task-specific adaptation makes it difficult to isolate the intrinsic transferability of the learned features, leaving it unclear whether any single brain-specific representation can generalize across fundamentally heterogeneous clinical endpoints.

This question is particularly challenging in brain MRI due to intrinsic heterogeneity across acquisition protocols, scanners, subject populations, and disease phenotypes~\cite{guan2021domain,yoon2024domain}. Moreover, downstream tasks impose distinct structural and semantic demands on the representation space. Tumor segmentation requires fine-grained boundary localization~\cite{menze2014multimodal,liu2023deep,allah2023edge}; neurodevelopmental and neurodegenerative classification depends on subtle global morphological patterns~\cite{frisoni2010clinical,feng2022deep,sarica2023explainability,zhang2022diagnosis}; temporal trajectory modeling and survival risk estimation rely on distributed structural biomarkers~\cite{cole2017predicting,bashyam2020mri,liew2018large,kickingereder2016radiomic,luckett2023predicting}. A unified representation must therefore be anatomically structured, disease-agnostic, and robust to distributional shifts~\cite{guan2021domain,yoon2024domain}. We hypothesize that this requires learning \textit{medical-specific
invariances} that suppress variation arising from acquisition physics
while preserving anatomically and clinically meaningful structure.
Because such invariances are absent from representations trained on
natural images, domain-specific pretraining on brain MRI data is
necessary even when the underlying SSL framework is shared.

Recent developments in natural image SSL offer a methodological
foundation for this goal. DINOv3~\cite{simeoni2025dinov3}, building upon the DINO family with refined training recipes and enhanced patch-level objectives, demonstrates that high-quality representations can support strong downstream performance under frozen feature extraction without task-specific adaptation~\cite{caron2021emerging}. This shift from fine-tuning-dependent representations to inherently sufficient ones motivates the exploration of analogous approaches in brain MRI. Initial evidence from adapting DINOv3 to CT-based organ and tumor segmentation~\cite{li2025meddinov3} further supports this direction.

\begin{figure}[t]
    \centering
    \includegraphics[width=\linewidth,height=0.9\textheight,keepaspectratio]{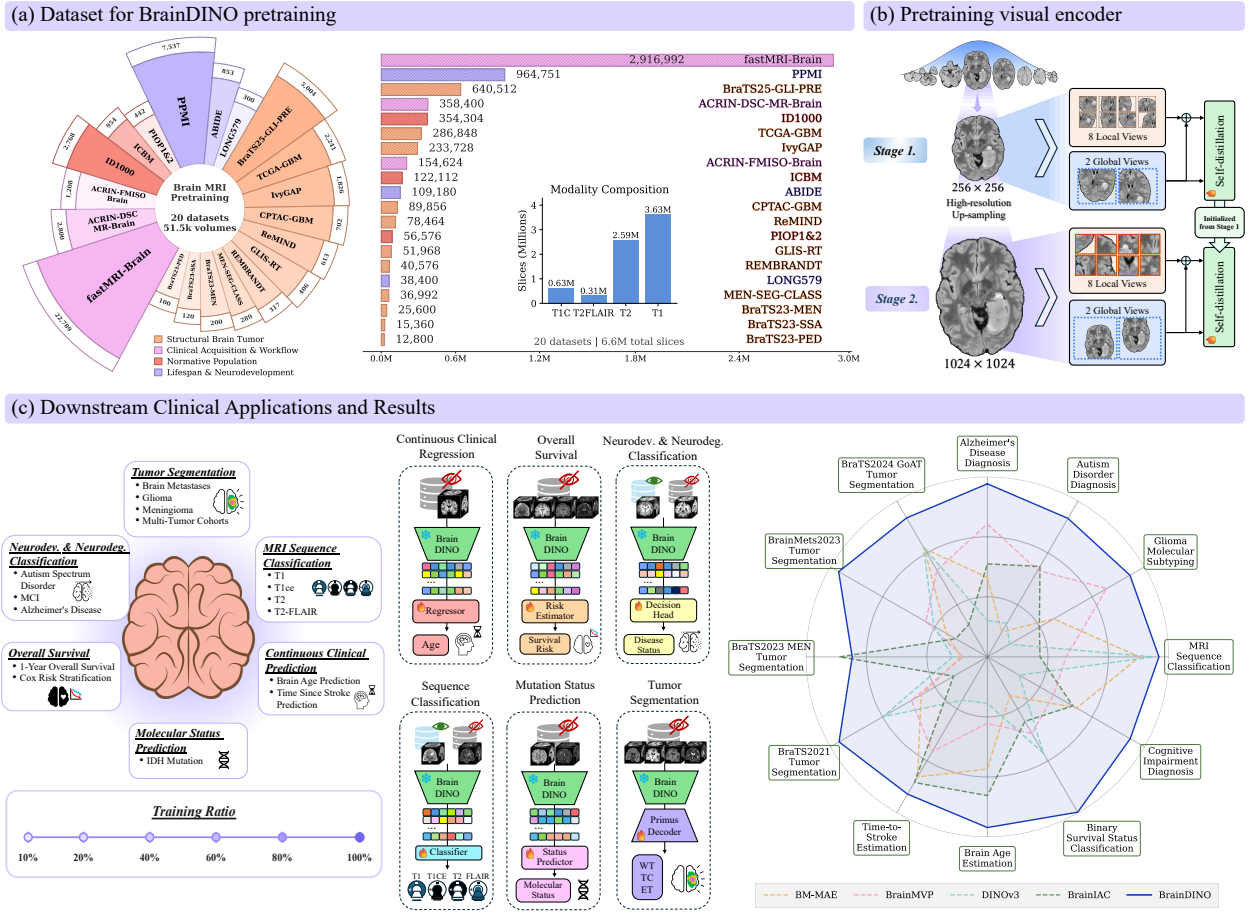}
\caption{
Overview of the proposed BrainDINO framework. 
(a) The pretraining corpus consists of large-scale and heterogeneous brain MRI datasets spanning normative populations, lifespan and neurodevelopment cohorts, structural brain tumor imaging, and clinical acquisition collections. Both volume-level composition and slice-level contributions are summarized, together with modality distribution including 1.97M T1, 2.60M T2, 0.88M T2-FLAIR, and 1.14M T1c slices.
(b) Slice-wise self-supervised pretraining is performed using a two-stage DINOv3-style self-distillation framework. Stage 1 learns global anatomical representations at standard resolution, and Stage 2 refines fine-grained structural features via high-resolution upsampling. Detailed training workflow is provided in Supplementary Fig.~1. 
(c) Frozen-backbone downstream evaluation across six clinical task families, including tumor segmentation, neurodevelopmental and neurodegenerative classification, molecular status prediction, MRI sequence classification, neuroanatomical trajectory modeling, and survival modeling.
Left: task families organized under varying labeled-data regimes (10\%–100\%) to assess data efficiency. 
Middle: unified model setup with a shared frozen encoder and lightweight task-specific heads. 
Right: performance comparison at full supervision (100\%), showing overall advantages of BrainDINO over alternative pretrained encoders across task categories.
}
    \label{fig:overview}
\end{figure}

In this work, we investigate whether large-scale brain-specific slice-wise SSL can yield such a unified and transferable representation. We adopt a DINOv3-style teacher--student self-distillation framework that jointly optimizes global semantic alignment via CLS-token distillation and local structural consistency via masked patch-token prediction~\cite{simeoni2025dinov3}, combined with multi-scale cropping to encourage scale-consistent and anatomically coherent feature learning~\cite{caron2021emerging} (Fig.~\ref{fig:overview}b). The model is pretrained on 6.6 million unlabeled axial slices collected from 20 heterogeneous brain MRI datasets (Fig.~\ref{fig:overview}a). The slice-wise formulation enables scalable training across diverse
cohorts while remaining agnostic to inter-slice continuity, an
assumption that may be unreliable in large-scale pretraining corpora
spanning heterogeneous diseases, varying slice thicknesses, and
diverse acquisition protocols.

To assess intrinsic representational generality, we evaluate the pretrained encoder under frozen-backbone adaptation using lightweight task-specific heads~\cite{chen2020simple,caron2021emerging}. We conduct systematic evaluation across six clinical task families, including neurodevelopmental and neurodegenerative classification, neuroanatomical trajectory modeling (brain age estimation and post-stroke temporal prediction), tumor segmentation, molecular status prediction, MRI sequence classification, and survival risk stratification (Fig.~\ref{fig:overview}c, left \& middle). By analyzing performance across varying labeled-data regimes, we examine not only downstream accuracy but also data efficiency under limited supervision. Across tasks and data regimes, BrainDINO consistently matches or exceeds representative baselines with the most pronounced gains emerging under label scarcity, demonstrating that large-scale slice-wise SSL is sufficient to produce a stable, transferable brain MRI representation without requiring volumetric pretraining or full-network fine-tuning (Fig.~\ref{fig:overview}c, right).

\section{Results}

We assessed the generality of BrainDINO across six clinical task families spanning tumor segmentation, neurological disorder classification, neuroanatomical trajectory modeling (brain age estimation and post-stroke temporal prediction), molecular status prediction, MRI sequence classification, and survival modeling. Tumor segmentation was evaluated on BraTS2021~\cite{baid2021rsna}, BraTS2023-Mets~\cite{moawad2024brain}, BraTS2023-MEN~\cite{labella2023asnr}, and BraTS2024-GoAT~\cite{bratsgoat2024}. Neurological disorder classification was conducted on ABIDE~\cite{di2014autism} (autism spectrum disorder), ADNI~\cite{petersen2010alzheimer} (Alzheimer's disease staging), OASIS~\cite{marcus2007open} (cognitive impairment), and PPMI~\cite{marek2011parkinson} (Parkinson's disease). Brain age regression used a combined IXI~\cite{ixi}, LONG579~\cite{wang2022longitudinal}, and Pixar~\cite{richardson2019mri} cohort, and post-stroke temporal prediction was evaluated on ATLAS~\cite{liew2018large, liew2022large}. Molecular status prediction targeted IDH mutation on UCSF-PDGM~\cite{ucsfpdgm}, MRI sequence classification used a curated cohort from BraTS2023~\cite{labella2023asnr, kazerooni2024brain,adewole2023brain}, and survival prediction was performed on UPENN-GBM~\cite{upenn-gbm}. Dataset details and train/test splits are provided in Supplementary Table~2.

We compare BrainDINO against four representative pretrained backbones: DINOv3 pretrained on natural images, and three MRI-specific self-supervised models--BrainMVP, BM-MAE, and BrainIAC. To assess data efficiency, all experiments were conducted under multiple labeled data availability regimes, ranging from 10\% to 100\% of the available training data. We further evaluated robustness under common MRI artifact perturbations.

\subsection{Tumor Segmentation}

We first evaluate the learned representation on brain tumor segmentation, a task that requires fine-grained spatial localization and accurate delineation of pathological regions. Experiments are conducted on four complementary benchmarks, including BraTS2021, BraTS2023-Mets, BraTS2023-MEN, and BraTS2024-GoAT, each emphasizing distinct tumor characteristics and generalization challenges.

BraTS2021 provides a structured benchmark for adult glioma segmentation with well-defined tumor subregions. BraTS2023-Mets focuses on brain metastases, where lesions are typically small, multiple, and spatially dispersed, stressing sensitivity to small targets and boundary precision. BraTS2023-MEN evaluates meningioma segmentation, testing transfer to a different tumor entity beyond glioma-specific patterns. BraTS2024-GoAT is designed to assess generalizability across tumor entities by spanning heterogeneous tumor populations and imaging conditions, thereby probing performance consistency across diverse segmentation tasks.

\subsubsection*{Quantitative Results}
Across all segmentation benchmarks, performance improves with increasing training data availability. Experiments were conducted at six labeled-data ratios (10\%, 20\%, 40\%, 60\%, 80\%, and 100\%), as summarized in Fig.~\ref{fig:Downstream_Results}(a). We highlighted representative results at 10\%, 60\%, and 100\% training data ratios in the main text, corresponding to low-, intermediate-, and full-data regimes (with concrete training set sizes reported per dataset). Segmentation accuracy was evaluated using the Dice similarity coefficient (Dice) for whole tumor (WT), tumor core (TC), and enhancing tumor (ET), reported as mean $\pm$ standard deviation. Complete per-ratio results for all four benchmarks are reported in Supplementary Table~7.

\paragraph{BraTS2021}
BrainDINO demonstrated consistent improvements over DINOv3 across all data 
regimes, with stable gains of approximately $+0.03$--$0.05$ Dice across all 
three subregions (Fig.~\ref{fig:Downstream_Results}(a)). Under limited 
supervision ($10\%$, $n=100$), BrainDINO achieved $0.916 \pm 0.015$ (WT), 
$0.889 \pm 0.018$ (TC), and $0.856 \pm 0.019$ (ET), exceeding DINOv3 
($0.885$, $0.850$, $0.812$; $p<0.05$). Performance improved progressively 
with additional labeled data, reaching $0.927 \pm 0.014$ (WT), 
$0.898 \pm 0.018$ (TC), and $0.872 \pm 0.019$ (ET) at full supervision 
($n=1000$), compared with $0.894$, $0.861$, and $0.818$ under DINOv3 
($p<0.05$).

\paragraph{BraTS2023-MEN}
BrainDINO maintained consistent advantages in WT segmentation across all data 
regimes, while TC and ET results revealed a more nuanced pattern 
(Fig.~\ref{fig:Downstream_Results}(a)). At $10\%$ training data ($n=80$), 
BrainDINO achieved WT Dice of $0.851 \pm 0.027$, compared with $0.754$ under 
DINOv3 ($p<0.05$). Under full supervision ($n=800$), WT reached 
$0.879 \pm 0.026$, exceeding DINOv3 ($0.786$) and BM-MAE ($0.795$) while 
remaining the highest among all backbones ($p<0.05$). For TC and ET, 
BrainDINO substantially improved over DINOv3 across all data regimes---at 
full data, TC and ET Dice reached $0.714 \pm 0.032$ and $0.729 \pm 0.032$,
compared with $0.537$ and $0.388$ under DINOv3. BrainDINO further remained the
strongest backbone on TC and ET across all data regimes, with every
MRI-specific SSL baseline---including BrainIAC ($0.469$ TC and $0.413$ ET at
$100\%$)---scoring significantly lower ($p<0.001$).

\paragraph{BraTS2023-Mets}
BrainDINO demonstrated the most pronounced advantages on this benchmark, 
where small and phenotypically variable lesions make limited-supervision 
performance particularly challenging (Fig.~\ref{fig:Downstream_Results}(a)). 
At $10\%$ training data ($n=19$), BrainDINO achieved $0.711 \pm 0.021$ (WT), 
$0.690 \pm 0.021$ (TC), and $0.666 \pm 0.020$ (ET), substantially exceeding 
all baselines---DINOv3 ($0.580$, $0.563$, $0.461$), BrainIAC ($0.537$, 
$0.524$, $0.481$), BM-MAE ($0.548$, $0.539$, $0.451$), and BrainMVP 
($0.560$, $0.545$, $0.462$; all $p<0.05$). Using $100\%$ of the training 
data ($n=190$), BrainDINO reached $0.778 \pm 0.019$ (WT), $0.760 \pm 0.019$ 
(TC), and $0.745 \pm 0.019$ (ET), maintaining the largest margins over all 
baselines ($p<0.05$), with narrowing confidence intervals indicating improved 
robustness at higher data availability.

\paragraph{BraTS2024-GoAT}
BrainDINO consistently achieved the highest Dice scores across all tumor 
subregions and data regimes on this heterogeneous multi-tumor benchmark 
(Fig.~\ref{fig:Downstream_Results}(a)). At $10\%$ training data ($n=108$), 
BrainDINO achieved $0.904 \pm 0.010$ (WT), $0.869 \pm 0.012$ (TC), and 
$0.828 \pm 0.012$ (ET), exceeding DINOv3 in WT and ET ($0.895$ and $0.814$; 
$p<0.05$) with a numerically higher but non-significant advantage in TC 
($0.864$). Reconstruction-based MRI-specific baselines ranged from 
$0.853$--$0.870$ (WT), $0.759$--$0.813$ (TC), and $0.770$--$0.785$ (ET), 
all significantly below BrainDINO ($p<0.05$). With $100\%$ of the training 
data ($n=1080$), BrainDINO reached $0.924 \pm 0.009$ (WT), 
$0.898 \pm 0.011$ (TC), and $0.852 \pm 0.012$ (ET), consistently 
outperforming all baselines ($p<0.05$).

\subsubsection*{Qualitative Analysis}
\begin{figure}[t]
    \centering
    \includegraphics[width=\linewidth,height=0.6\textheight,keepaspectratio]{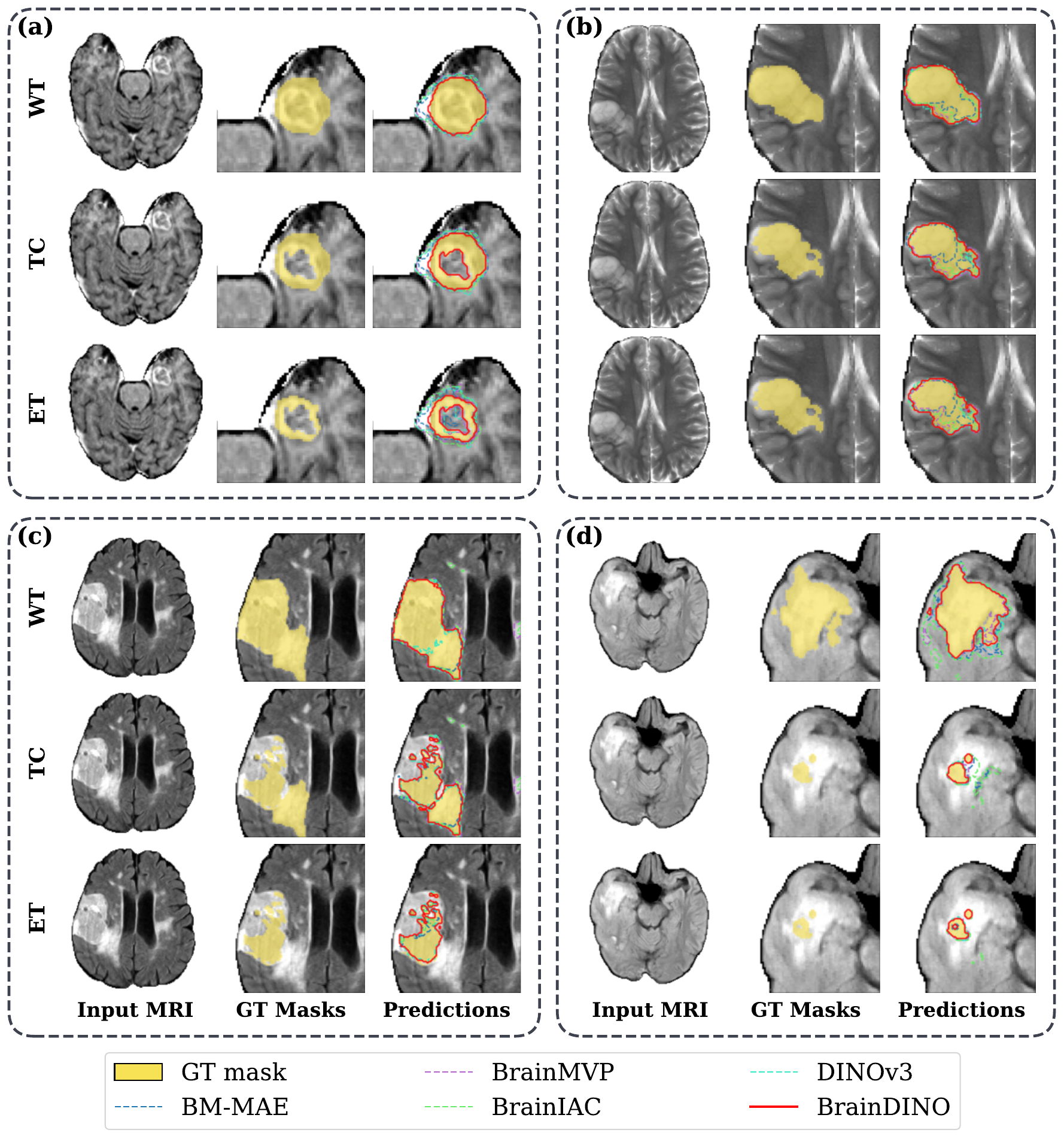}
    \caption{Cross-dataset qualitative comparison of tumor segmentation under full-data training. Representative segmentation results on BraTS2023-Mets and BraTS2021 benchmarks using models fine-tuned with 100\% of labeled training data. For each dataset, we show the input MRI slice, ground-truth (GT) mask (yellow), and predictions from different pretrained backbones (colored contours). Three tumor subregions are visualized: whole tumor (WT), tumor core (TC), and enhancing tumor (ET).}
    \label{fig:Qualitative_Comparison}
\end{figure}

Fig.~\ref{fig:Qualitative_Comparison} illustrates representative cross-dataset segmentation examples. Across both BraTS2023-Mets and BraTS2021, BrainDINO (red contour) demonstrates closer spatial alignment with the ground-truth masks, with improved boundary conformity and reduced over-segmentation compared to alternative pretrained backbones. Competing models tend to produce enlarged contours or fragmented delineations, particularly along irregular tumor margins and heterogeneous core regions. These qualitative observations are consistent with the quantitative trends and support improved cross-dataset generalization under fully supervised fine-tuning.

\subsection{Neurodevelopmental and Neurodegenerative Classification}

We next assessed whether the learned representation captures disease-oriented semantic information through neurodevelopmental and neurodegenerative classification. Experiments were conducted on four structural T1-weighted MRI benchmarks including ABIDE, ADNI, OASIS, and PPMI spanning distinct dataset scales and diagnostic complexity.

ABIDE is formulated as 
a binary classification task distinguishing autism spectrum disorder (ASD) from 
healthy controls, comprising 956 training subjects and 240 test subjects. The
ABIDE training cohort partially overlapped with the pretraining corpus, while
test subjects were strictly held out. ADNI represents a substantially 
more challenging multi-class setting involving cognitively normal (CN), mild cognitive 
impairment (MCI), and Alzheimer's disease (AD). It comprises 2{,}182 T1-weighted scans
from 382 subjects (CN/MCI/AD $=$ 748/981/453 scans) and is evaluated under a strict
subject-disjoint protocol (no subject shared between train and test),
with no ADNI subjects used during pretraining.
OASIS is a smaller independent dementia cohort consisting of 195 training
subjects and 40 test subjects, serving as a low-sample evaluation scenario.
PPMI is a binary classification task distinguishing Parkinson's disease (PD)
from healthy controls, comprising 1{,}152 training subjects and 289 test subjects.
Classification performance was evaluated using macro-AUC under training data ratios 
ranging from 10\% to 100\%. Complete per-ratio macro-AUC for all four cohorts is reported in Supplementary Table~8. 

\begin{figure}[H]
    \centering  \includegraphics[width=\linewidth,height=0.85\textheight,keepaspectratio]{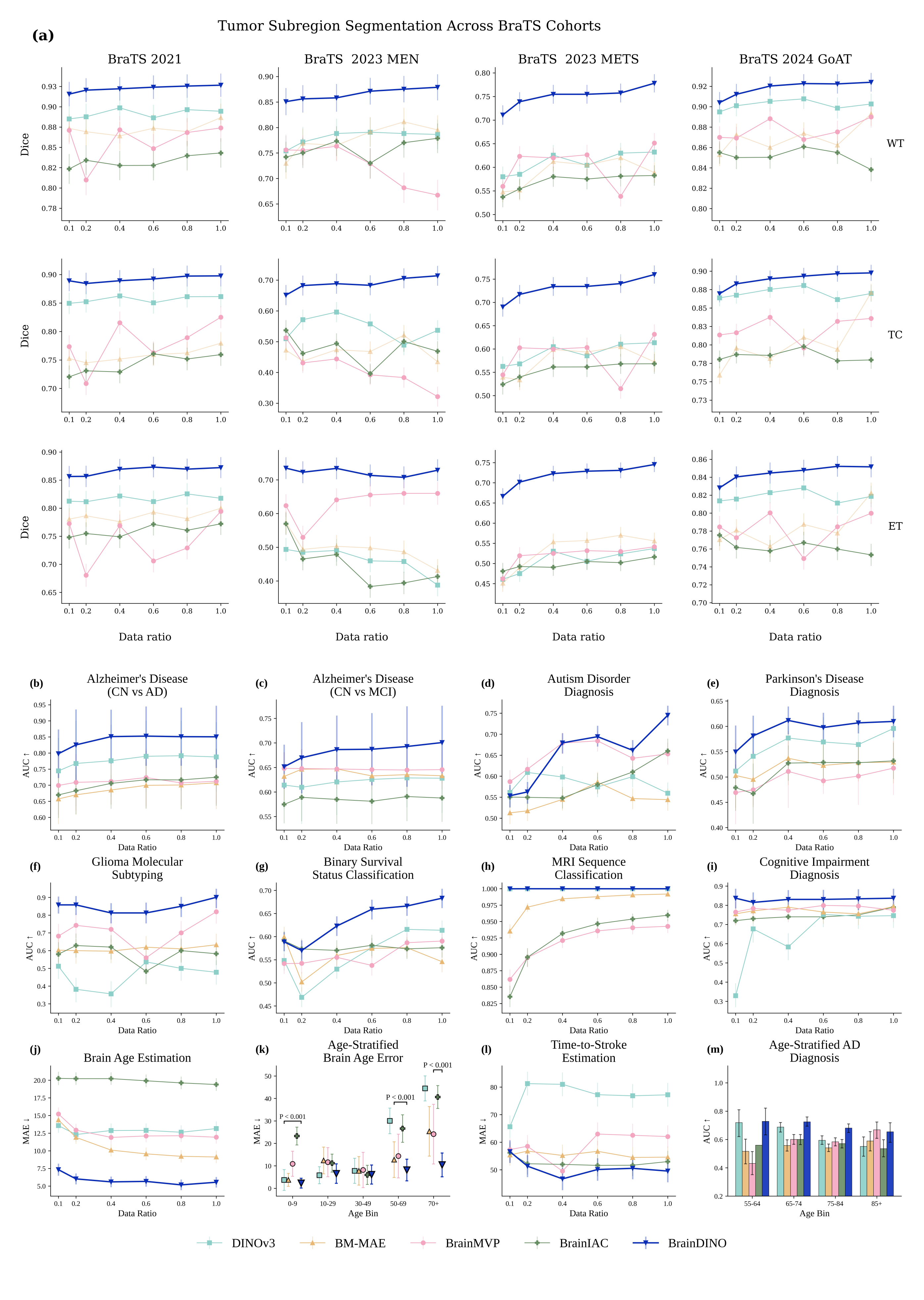}
\caption{Summary of downstream evaluation across segmentation and
neuroimaging tasks. ADNI uses a strict subject-level protocol
(subject-disjoint cross-validation; no subject shared between train and test) to
avoid longitudinal leakage; the figure shows the two binary contrasts
(CN-vs-AD, CN-vs-MCI) and the age-stratified analysis, with the full
3-class CN/MCI/AD result in Supplementary Table~8.
\textbf{(a)}~Tumor subregion segmentation (Dice) across four BraTS
cohorts vs.\ labeled-data ratio.
\textbf{(b)}~ADNI CN-vs-AD diagnosis (macro-AUC).
\textbf{(c)}~ADNI CN-vs-MCI early detection (macro-AUC).
\textbf{(d)}~ASD classification on ABIDE.
\textbf{(e)}~Parkinson's diagnosis on PPMI.
\textbf{(f)}~IDH mutation prediction on UCSF-PDGM.
\textbf{(g)}~Overall-survival classification on UPENN-GBM.
\textbf{(h)}~MRI sequence classification on BraTS2023.
\textbf{(i)}~Cognitive impairment diagnosis on OASIS.
\textbf{(j)}~Brain age estimation (MAE) on IXI\,+\,LONG579\,+\,Pixar.
\textbf{(k)}~Age-stratified brain age error at 100\% supervision.
\textbf{(l)}~Post-stroke temporal prediction (MAE) on ATLAS.
\textbf{(m)}~Age-stratified 3-class AD staging on ADNI (100\%
supervision; pooled subject-disjoint test predictions across splits).
Significance: $*\,p<0.05$, $**\,p<0.01$, $***\,p<0.001$ (panels~k, m);
full significance in the Supplementary Materials.}

    \label{fig:Downstream_Results}
\end{figure}

\paragraph{ABIDE}
As shown in Fig.~\ref{fig:Downstream_Results}(d), performance differences 
on ABIDE were modest at low data regimes, with no statistically significant 
differences among backbones at $10\%$ or $20\%$ supervision. Statistically 
significant improvements emerged from $40\%$ training data onward, where 
BrainDINO reached $0.679$, outperforming BrainIAC ($0.548$, $p<0.05$), 
BM-MAE ($0.545$, $p<0.05$), and DINOv3 ($0.598$, $p<0.05$). Under full 
supervision ($100\%$), BrainDINO attained $0.745$, exceeding BrainIAC 
($0.660$, $p<0.05$), BM-MAE ($0.544$, $p<0.05$), and DINOv3 ($0.559$, 
$p<0.05$), while remaining comparable to BrainMVP ($0.653$). These results 
suggest that moderate supervision is required to unlock statistically robust 
advantages on this coarse-grained binary task.

\paragraph{ADNI}
ADNI was evaluated under a \emph{subject-level} protocol: no scan from a
given subject appears in both train and test, eliminating the
longitudinal-visit leak that inflates scan-level splits. The main figure
reports two clinically motivated binary formulations of increasing
difficulty---CN-vs-AD (the standard Alzheimer's diagnostic comparison)
and CN-vs-MCI (early-detection screening: distinguishing cognitively
normal individuals from those with subtle prodromal impairment)---while
the full 3-class CN/MCI/AD results are reported in Supplementary Table~8. All ADNI metrics are
macro-AUC averaged over $5$ subject-disjoint splits.

On CN-vs-AD (Fig.~\ref{fig:Downstream_Results}(b)), BrainDINO reached $0.850$
[$0.754$, $0.947$] at full supervision, outperforming BrainMVP ($0.712$,
$p=0.003$), BM-MAE ($0.707$, $p=0.003$), BrainIAC ($0.725$, $p=0.022$), and
DINOv3 ($0.788$, $p=0.070$). The advantage was already established under
heavy label scarcity: at $10\%$ supervision BrainDINO achieved $0.798$
versus BM-MAE $0.657$ ($p=0.005$), BrainIAC $0.670$ ($p=0.029$), and
BrainMVP $0.699$ ($p=0.045$).

On the more challenging CN-vs-MCI early-detection distinction
(Fig.~\ref{fig:Downstream_Results}(c)), BrainDINO again ranked first at
every supervision level, reaching $0.701$ [$0.608$, $0.794$] at full data,
exceeding DINOv3 ($0.628$, $p=0.053$), BrainMVP ($0.646$, $p=0.144$),
BM-MAE ($0.633$, $p=0.122$), and BrainIAC ($0.588$, $p=0.001$). Absolute
margins here are smaller than on CN-vs-AD, consistent with the inherent
difficulty of detecting the subtle structural changes that distinguish
normal aging from prodromal cognitive impairment.

Age-stratified analysis under full supervision
(Fig.~\ref{fig:Downstream_Results}(m)) further showed BrainDINO leading the
$55$--$64$, $65$--$74$, and $75$--$84$ strata (macro-AUC $0.728$, $0.725$,
$0.681$); in the small $85$+ stratum BrainMVP was numerically higher
($0.668$ vs $0.654$; complete per-stratum values are reported in Supplementary Table~15).

\paragraph{OASIS}
On the smaller OASIS cohort, BrainDINO consistently achieved the highest 
macro-AUC across all training data regimes 
(Fig.~\ref{fig:Downstream_Results}(i)). At $10\%$ supervision, BrainDINO 
reached $0.837$, significantly outperforming DINOv3 ($0.329$, $p<0.05$) 
and BrainIAC ($0.687$, $p<0.05$), while maintaining numerical advantages 
over BrainMVP ($0.765$) and BM-MAE ($0.755$). Performance remained stable 
across all data regimes ($0.815$--$0.837$), whereas competing models showed 
greater variability. Due to the small test set ($n=40$), BM-MAE and BrainMVP 
comparisons did not reach statistical significance across most ratios, 
highlighting variability inherent to small-sample evaluation.

\paragraph{PPMI}
We evaluated PD versus healthy-control classification on PPMI, comprising 1158 PD 
and 283 control subjects with baseline T1-weighted MRI, a neurodegenerative cohort 
absent from all other classification benchmarks. As shown in Fig.~\ref{fig:Downstream_Results}(e), 
BrainDINO achieved the highest macro-AUC across all training data ratios. At $10\%$ 
supervision, BrainDINO reached $0.549$~[$0.485$, $0.614$], exceeding DINOv3 
($0.512$), BM-MAE ($0.503$), BrainIAC ($0.479$), and BrainMVP ($0.469$). Under full 
supervision ($100\%$), BrainDINO attained $0.610$~[$0.571$, $0.648$], maintaining a 
consistent advantage over DINOv3 ($0.596$), BrainIAC ($0.532$), BM-MAE ($0.529$), 
and BrainMVP ($0.518$). Although absolute performance remained modest across all 
backbones, consistent with the limited structural correlates of PD on T1-weighted 
MRI, the uniform BrainDINO advantage indicates that brain-specific pretraining 
yields a more discriminative representation even for this fine-grained, 
low-signal task.



\subsection{Neuroanatomical Trajectory Modeling}

We next evaluated whether the learned representation captures continuous neuroanatomical trajectories through brain age estimation and post-stroke temporal prediction.
\paragraph{Brain Age Estimation}
BrainDINO consistently achieved the lowest mean absolute error (MAE, years) across all training data 
ratios on the combined IXI, LONG579, and Pixar cohort 
(Fig.~\ref{fig:Downstream_Results}(j)), significantly outperforming all 
baselines at every supervision level ($p<0.05$). Under extreme label 
limitation ($10\%$, $n=144$), BrainDINO attained an MAE of  $7.33 \pm 0.85$ years, 
compared with DINOv3 ($13.58 \pm 1.02$), BrainMVP ($15.22 \pm 0.92$),
and BrainIAC ($20.26 \pm 0.94$), corresponding to an absolute reduction
of $6$--$13$ years. Notably, BrainDINO with only $20\%$ labeled data 
($6.02 \pm 0.77$ years) already surpassed all baselines trained with 
$100\%$ data, demonstrating strong data efficiency. Under full supervision 
($100\%$, $n=1440$), BrainDINO maintained the lowest MAE at  
$5.54 \pm 0.75$ years, while the strongest baseline (BM-MAE) reached
an MAE of $9.12 \pm 0.88$ years, with BrainIAC the weakest at $19.40 \pm 0.89$ years ($p<0.05$). Per-ratio values are tabulated in Supplementary Table~9.

Age-stratified MAE under full supervision 
(Fig.~\ref{fig:Downstream_Results}(k)) revealed a nuanced lifespan 
profile (Supplementary Table~16). 
BrainDINO achieved the strongest performance in the youngest ($0$--$9$: 
$2.31$ years vs.\ DINOv3 $3.70$, $p<0.05$) and oldest ($70+$: $10.45$ 
years vs.\ BrainMVP $24.13$, $p<0.05$) age groups, as well as in the 
$50$--$69$ bin ($8.16$ vs.\ BM-MAE $12.74$, $p<0.05$). In the mid-life 
bins ($10$--$49$), differences were not statistically significant, with 
DINOv3 and BrainIAC achieving marginally lower errors in the $10$--$29$ 
and $30$--$49$ groups respectively. These findings suggest that 
brain-specific pretraining yields the greatest gains at the extremes of 
the age distribution, where normative aging signatures are most pronounced.

\paragraph{Post-Stroke Temporal Prediction}
We further evaluated pathological temporal modeling on the ATLAS dataset 
by predicting Days Post Stroke (DPS), a task assessing lesion-driven 
structural evolution rather than normative aging, measured by MAE (days)
(Fig.~\ref{fig:Downstream_Results}(l)). At $10\%$ supervision ($n=21$), 
BrainDINO achieved an MAE of $56.6 \pm 4.1$ days, comparable to BM-MAE 
($55.4$) and BrainIAC ($56.2$), with no statistically significant 
differences among MRI-specific models. From $20\%$ supervision onward, 
BrainDINO consistently outperformed DINOv3 by a substantial margin 
($51.3$ vs.\ $81.3$ days at $20\%$; $49.5$ vs.\ $77.2$ days at $100\%$; 
all $p<0.05$). The strongest performance was observed at $40\%$ supervision 
(MAE $46.7 \pm 4.1$ days). MRI-specific baselines BrainIAC and BM-MAE 
remained numerically close to BrainDINO across all regimes without 
reaching statistical significance, reflecting the high-variance nature of 
this small-sample task. Full results across all data ratios appear in Supplementary Table~11.

\subsection{Overall Survival Classification}

\paragraph{Binary Classification}
Binary overall survival classification was evaluated on the UPENN-GBM 
cohort using macro-AUC, with 121 subjects in the held-out test set
(Fig.~\ref{fig:Downstream_Results}(g)). At $10\%$ supervision ($n=48$), 
BrainDINO achieved a macro-AUC of $0.589$ [0.568, 0.610], comparable to 
BrainIAC ($0.591$) and BM-MAE ($0.599$), and higher than BrainMVP 
($0.542$) and DINOv3 ($0.548$), with overlapping confidence intervals 
across all comparisons. Performance improved progressively with additional 
supervision across all backbones. Under full supervision ($100\%$, 
$n=482$), BrainDINO achieved the highest macro-AUC of $0.683$ 
[0.662, 0.704], exceeding BM-MAE ($0.546$, $p<0.05$), while differences 
relative to BrainIAC ($0.576$), BrainMVP ($0.590$), and DINOv3 ($0.614$) 
did not reach statistical significance. Supplementary Table~13 reports the complete per-ratio breakdown.


\begin{figure}[t]
    \centering
    \includegraphics[width=\linewidth,height=1\textheight,keepaspectratio]{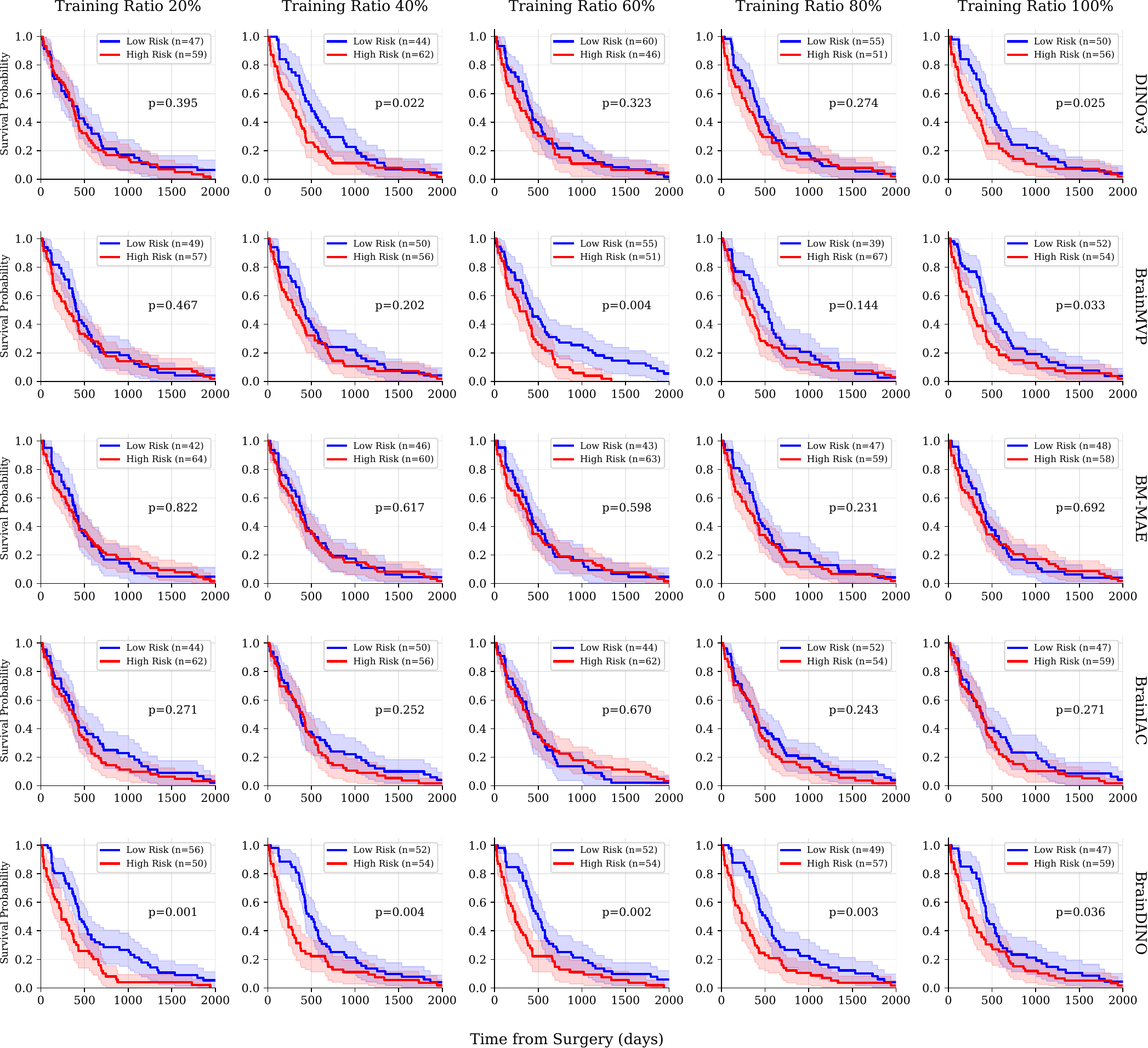}
    \caption{Kaplan-Meier survival analysis across training data ratios on UPENN-GBM. Survival curves for low- and high-risk patient groups stratified by predicted risk scores from five pretrained backbones (BrainDINO, DINOv3, BrainMVP, BrainIAC, and BM-MAE) under five training data availability settings (20\%, 40\%, 60\%, 80\%, and 100\%). Statistical significance is assessed using log-rank tests. For visualization clarity, the time axis is truncated to 2,000 days.}
    \label{fig:km_ratio_sweep}
\end{figure}

\paragraph{Kaplan--Meier Risk Stratification}
We next evaluated survival modeling using Kaplan-Meier risk stratification 
based on median risk scores estimated from the training set. BrainDINO 
consistently demonstrated statistically significant separation between 
high- and low-risk groups across all training data ratios ($p<0.05$ at 
all ratios; Fig.~\ref{fig:km_ratio_sweep}). Clear divergence of survival 
trajectories was observed even at low data availability, with low-risk 
median survival of $439$ days versus $235$ days for high-risk patients 
at $20\%$ supervision, and remained stable as supervision increased. In 
contrast, alternative backbones showed inconsistent or non-significant 
risk separation---DINOv3 achieved significance only at $40\%$ and $100\%$, 
while BM-MAE and BrainIAC failed to reach significance at any ratio. 
These results indicate that BrainDINO supports robust survival risk 
stratification across varying levels of supervision; see Supplementary Table~14 for all data ratios.

\subsection{Mutation Detection}
We evaluated molecular-level inference using IDH mutation status prediction 
on the UCSF-PDGM cohort. After filtering for subjects with both T1CE and 
FLAIR modalities available, 360 subjects were used for training and 92 
subjects for independent testing (73 IDH-wildtype and 19 non-wildtype). 
Performance was assessed using AUC across varying training data ratios 
(Fig.~\ref{fig:Downstream_Results}(f)).

At $10\%$ training data, BrainDINO achieved an AUC of $0.858$ 
[0.809, 0.905], significantly exceeding BM-MAE ($0.603$, $p<0.05$), 
BrainIAC ($0.580$, $p<0.05$), DINOv3 ($0.512$, $p<0.05$), and BrainMVP 
($0.682$, $p<0.05$). At intermediate supervision ($40\%$--$60\%$), 
BrainDINO maintained AUC of $0.813$, significantly outperforming BM-MAE, 
BrainIAC, and DINOv3 ($p<0.05$), while differences relative to BrainMVP 
were not statistically significant at these ratios. Under full supervision 
($100\%$), BrainDINO achieved its highest AUC of $0.901$ [0.840, 0.949], 
significantly outperforming BM-MAE ($0.633$, $p<0.05$), BrainIAC 
($0.583$, $p<0.05$), and DINOv3 ($0.478$, $p<0.05$), while the difference 
relative to BrainMVP ($0.819$) did not reach statistical significance. 
Across all supervision regimes, BrainDINO consistently achieved the highest 
AUC, with confidence intervals shifted toward higher performance relative 
to all baselines. Complete results are provided in Supplementary 
Table~10.

\subsection{MRI Sequence Classification}
We further evaluated whether the learned representation encodes 
acquisition-level semantics through MRI sequence classification, a task 
probing sensitivity to protocol and contrast differences independent of 
pathology. Experiments were conducted on a curated cohort aggregated from BraTS2023-MEN,
BraTS2023-PED, and BraTS-Africa (BraTS2023-SSA), comprising four MRI sequences
(T1, T1c, T2, and T2-FLAIR; 3{,}853 training and 971 test volumes). To place all
backbones on an equal footing, every encoder was evaluated under a matched
frozen-feature protocol: per-scan pooled features were extracted once from each
frozen backbone and classified by a lightweight head trained with identical
hyperparameters.

Both 2D slice-based ViT encoders---BrainDINO and DINOv3---saturated this
acquisition-level task, reaching near-perfect macro-AUC ($\approx 1.000$) at
every supervision level and remaining statistically indistinguishable from each
other (Fig.~\ref{fig:Downstream_Results}(h)). Both substantially and
significantly outperformed the 3D volumetric baselines at all data ratios, with
the largest margins in the low-data regime: at $10\%$ supervision, BrainDINO
reached a macro-AUC of $1.000$, versus BM-MAE ($0.935$), BrainMVP ($0.862$), and
BrainIAC ($0.835$; all $p<0.001$). The volumetric baselines improved with more
labels but remained well below the 2D encoders even at full supervision (BM-MAE
$0.992$, BrainIAC $0.960$, BrainMVP $0.943$ at $100\%$). These results indicate
that slice-based ViT representations are highly sensitive to acquisition-level
contrast differences; on this particular task brain-specific pretraining confers
no measurable advantage over a strong natural-image-pretrained ViT, while both
clearly exceed the volumetric brain encoders. Complete results are provided in
Supplementary Table~12.

\begin{figure}[ht]
    \centering
    \includegraphics[width=\linewidth,height=10\textheight,keepaspectratio]{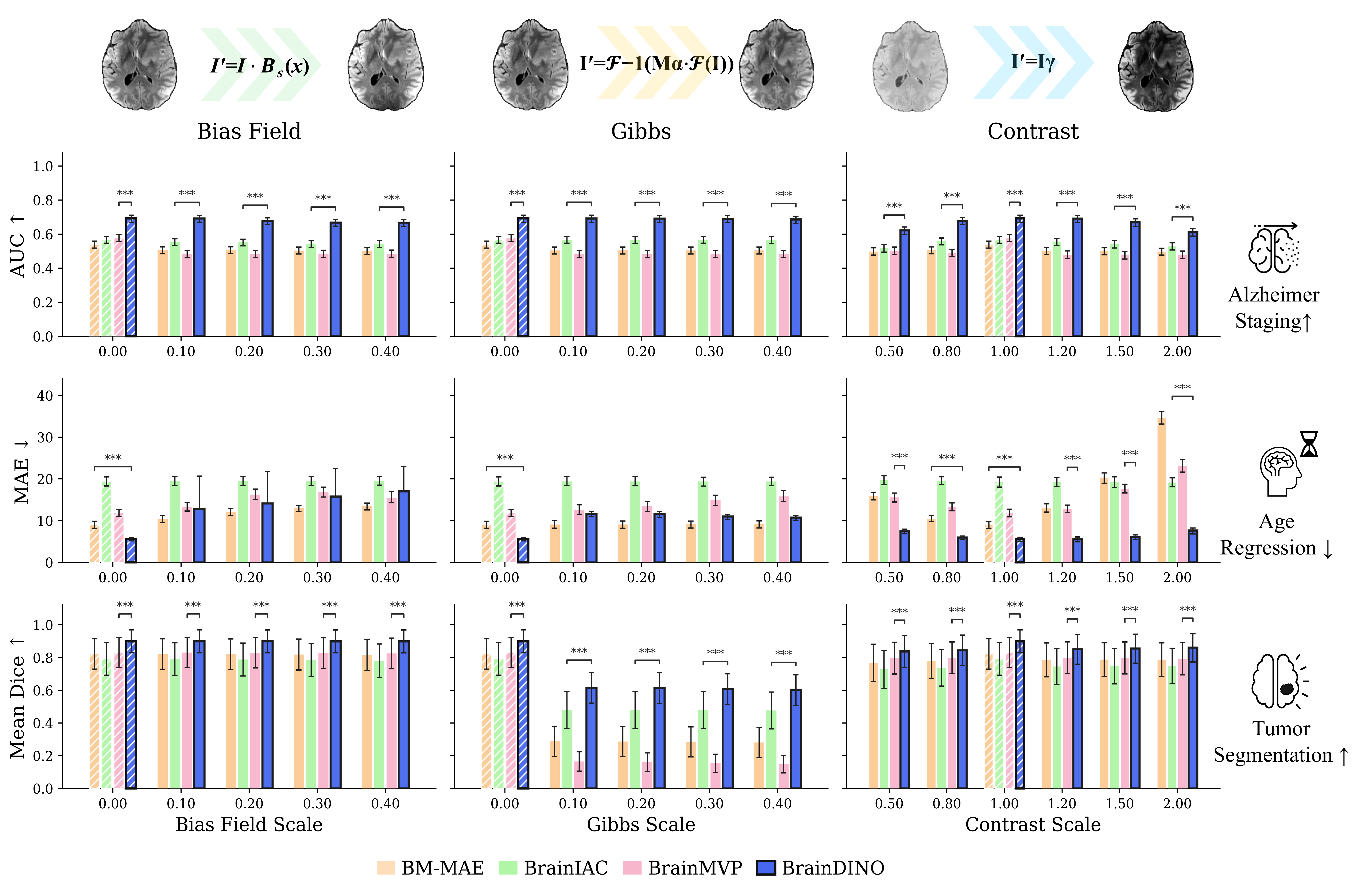}
\caption{Test-time perturbation analysis across ADNI Alzheimer's disease staging (CN/MCI/AD; macro-AUC), brain age regression (MAE), and BraTS2021 tumor segmentation (Mean Dice for WT/TC/ET) under three representative perturbations (Bias Field, Gibbs, and Contrast). Grouped bars show performance across increasing perturbation severity levels for four pretrained encoders, with error bars indicating uncertainty estimates. BrainDINO achieves the strongest clean baseline and remains robust across most perturbation settings, although degradation patterns are task-dependent, with segmentation showing pronounced sensitivity to Gibbs artifacts. Complete numerical results and pairwise statistics are reported in Supplementary Table~18.}

    \label{fig:Perturbation}
\end{figure}

\subsection{Frozen-Backbone Adaptation versus Full Fine-Tuning}
To assess whether the frozen-backbone protocol substantially limits downstream
performance, we compared frozen adaptation with full fine-tuning on four
representative conditions spanning task difficulty and supervision type: two
ADNI binary tasks under the same subject-level protocol used elsewhere in the
paper---CN-vs-AD (Alzheimer's disease diagnosis) and CN-vs-MCI
(early detection)---brain age regression, and BraTS2021
tumor segmentation (Supplementary Table~6). On CN-vs-AD, full fine-tuning is
not statistically distinguishable from frozen adaptation at either supervision
level (0.798 vs.\ 0.801 at 10\% data; 0.850 vs.\ 0.872 at 100\%; Full~FT
falls within the Frozen cross-validation confidence interval in both cases). On the more
challenging CN-vs-MCI early-detection task the frozen encoder remains on par
with full fine-tuning: the two are statistically tied at full supervision
(0.701 vs.\ 0.743; Full~FT within the Frozen cross-validation confidence interval), while
at 10\% data the frozen encoder is modestly \emph{better}
($\Delta = +0.073$ AUC; 0.652 vs.\ 0.579, Full~FT below the Frozen CI), as
full-network adaptation overfits the limited subject-disjoint sample. On BraTS2021, the frozen encoder matches full
fine-tuning at 10\% supervision ($|\Delta| \leq 0.001$ Dice) and retains a
small but significant advantage on WT and ET at full supervision. Only
continuous brain age regression shows a clear, consistent benefit from full
fine-tuning (MAE 7.33$\rightarrow$6.24 at 10\%; 5.54$\rightarrow$3.29 at
100\%). Frozen-backbone adaptation thus incurs minimal performance cost---and on harder, lower-signal classification
tasks even outperforms full fine-tuning---while updating only
0.6\%--22.6\% of model parameters, supporting its use as a
parameter-efficient and practically viable adaptation strategy.

\subsection{Perturbation Analysis}

We further examine the stability of the learned representations under common MRI perturbations to assess generalization under distributional shifts. Perturbation robustness is not treated as a primary contribution of this work; rather, this analysis characterizes how pretrained backbones respond when image statistics are systematically altered at test time. Following a standardized test-time perturbation protocol, we consider three clinically relevant perturbation families: global intensity rescaling via gamma correction (contrast perturbation), Gibbs artifact (k-space truncation), and bias field inhomogeneity. Perturbation severity was progressively increased within predefined ranges, while all encoders remain frozen and no training-time augmentation or adaptation is introduced. Experiments were conducted across three downstream tasks with distinct supervision characteristics: three-class neurodegenerative classification on ADNI, brain age regression, and brain tumor segmentation on BraTS2021 (Fig.~\ref{fig:Perturbation}).

On ADNI 3-class staging, evaluated under a strict subject-level protocol (no scan from a given subject appears in both train and test, with predictions pooled across subject-disjoint test splits), BrainDINO achieved the strongest clean baseline performance (macro-AUC $0.69$), exceeding its closest competitor BM-MAE ($0.64$) as well as BrainMVP ($0.62$) and BrainIAC ($0.59$). As perturbation severity increased, all models degraded, yet BrainDINO retained the highest macro-AUC across every perturbation family. Under severe contrast perturbation ($\gamma=2.0$), BrainDINO retained a macro-AUC of $0.63$, compared to $0.55$ for BrainIAC, $0.51$ for BM-MAE, and $0.42$ for BrainMVP. For Gibbs perturbation at strength $0.4$, BrainDINO was essentially unaffected ($0.70$), while competing models remained at $0.59$ (BrainIAC), $0.58$ (BM-MAE), and $0.41$ (BrainMVP). Under bias field perturbation at strength $0.4$, BrainDINO reached $0.71$, remaining substantially above all baselines. BrainDINO consistently maintained superior absolute performance across all perturbation strengths (see Supplementary Table~18), indicating stable relative generalization under distributional shifts.

For brain age regression, BrainDINO achieves the lowest clean MAE ($5.5$ years), markedly outperforming BM-MAE ($9.1$), BrainMVP ($11.9$), and BrainIAC ($19.4$). Under contrast perturbation, BrainDINO remains comparatively stable and preserves the lowest error across the full severity range (MAE $7.6$ at $\gamma=2.0$), while BM-MAE exhibits substantial error amplification (MAE $34.6$). In contrast, bias field and Gibbs perturbations introduce stronger degradation for all models. BrainDINO demonstrates noticeable error increases under severe bias field and Gibbs perturbations, reflecting the task’s sensitivity to global intensity gradients and high-frequency cortical structure. Despite this degradation, BrainDINO remains competitive and avoids catastrophic error escalation observed in several baselines under extreme contrast perturbation.

On brain tumor segmentation (BraTS2021), BrainDINO achieved the highest performance across all subregions  under clean conditions. Bias field perturbation produces minimal degradation across all models, with BrainDINO exhibiting nearly unchanged Dice scores even at the highest severity. Moderate contrast perturbation leads to small performance reductions, while relative ranking remains consistent. In contrast, Gibbs perturbation induced substantial degradation for all backbones, particularly for WT segmentation, reflecting the strong dependence of boundary delineation on high-frequency structural information. Although absolute Dice scores decrease sharply under severe Gibbs corruption, BrainDINO maintains competitive performance relative to alternative encoders. These findings indicate that segmentation robustness is primarily limited by perturbation type rather than encoder architecture.

\begin{figure}[ht]
    \centering
    \includegraphics[width=\linewidth]{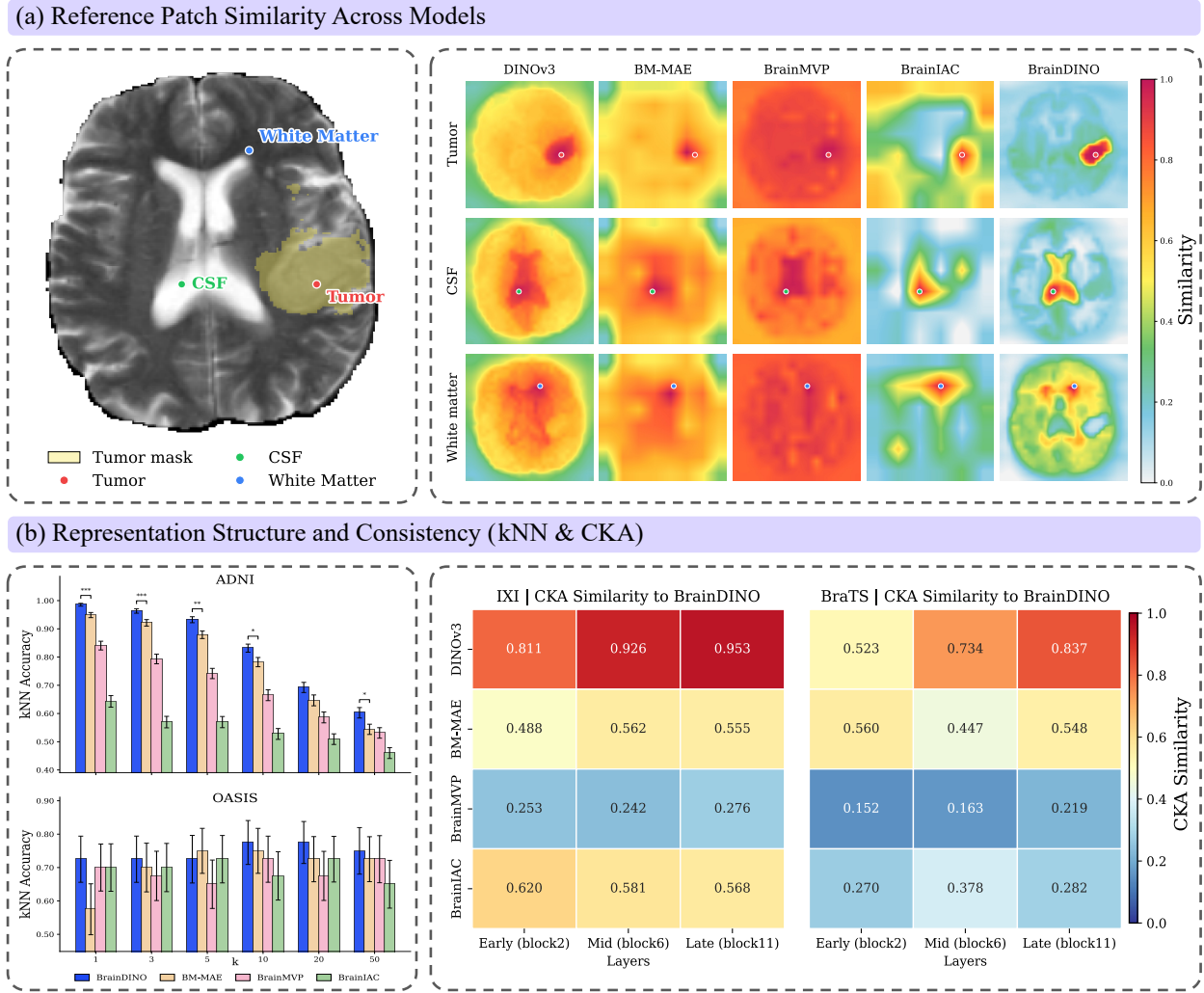}
    \caption{\textbf{Representation structure and downstream discriminability of frozen pretrained encoders.}
    \textbf{(a)}~Reference-point patch similarity across models. For a representative BraTS subject, similarity maps are computed from three anatomically distinct reference patches (tumor core, CSF, white matter) using frozen patch-token features from five pretrained backbones. BrainDINO produces spatially selective similarity distributions that closely correspond to anatomical compartments, whereas natural-image-pretrained DINOv3 yields more diffuse patterns.
    \textbf{(b)}~Representation structure and consistency. \emph{Left:} $k$NN classification accuracy on ADNI (top) and OASIS (bottom) using frozen features from four MRI-specific SSL backbones across neighborhood sizes $k \in \{1, 3, 5, 10, 20, 50\}$. Error bars denote standard deviations estimated via stratified bootstrap resampling of the test set (2{,}000 iterations). BrainDINO achieves the highest or tied-highest accuracy on ADNI, with accuracy increasing across $k$; on OASIS, numerical advantages are present but confidence intervals are wider due to the small test set ($n=40$). \emph{Right:} Layer-wise centered kernel alignment (CKA) between frozen BrainDINO representations at early (block\,2), mid (block\,6), and late (block\,11) layers and the natural-image-pretrained (DINOv3) and brain-MRI-pretrained (BM-MAE, BrainMVP, BrainIAC) backbones, computed on normative (IXI) and pathological (BraTS, 250 subjects) cohorts. On IXI, BrainDINO remains strongly aligned with DINOv3 across depth. On BraTS, alignment with DINOv3 increases with depth while alternative MRI-specific backbones exhibit lower and less stable similarity in deeper layers, indicating distinct representational geometries under pathological distribution shift.}
    \label{fig:rep_structure}
\end{figure}

\subsection{Representation-Level Analysis}
To further examine the intrinsic properties of the learned representation beyond task-specific performance, we conducted representation-level analyses focusing on feature separability, spatial organization, and cross-model representational consistency under frozen encoders.

Since $k$NN classification requires no learned parameters, its performance 
directly reflects the intrinsic class structure of the frozen feature 
space---free from any confound introduced by task-specific optimization. 
To assess whether the learned representations are inherently class-separable 
prior to any downstream adaptation, we applied nonparametric $k$NN 
probing~\cite{caron2021emerging,oquab2023dinov2} on frozen CLS-token 
features for binary dementia classification on ADNI (CN vs.\ AD) and
OASIS across neighborhood sizes
$k \in \{1, 3, 5, 10, 20, 50\}$. To avoid leakage from ADNI's repeated
longitudinal visits, the ADNI reference and query sets were partitioned
at the subject level, with no subject appearing in both. As shown in
Fig.~\ref{fig:rep_structure}(b, left), BrainDINO achieved
the highest or tied-highest accuracy at every $k$ on ADNI, with accuracy
increasing from $0.64$ ($k{=}1$) to $0.75$ ($k{=}50$); its margin over the
competing MRI-specific backbones widened at larger neighborhood sizes and
reached statistical significance against all three backbones at $k{=}50$
($p<0.05$, exact McNemar). On OASIS, BrainDINO also demonstrated consistent
numerical advantages, although no pairwise comparison reached statistical 
significance due to the smaller test set ($n=40$). These results indicate 
that the learned representation exhibits stronger intrinsic class 
separability compared to competing backbones prior to any task-specific 
adaptation (full numerical results in Supplementary 
Table~17).

To further characterize the spatial and structural organization of the learned feature space, we computed reference-point similarity maps using frozen patch-token embeddings, measuring similarity between selected anatomical reference patches (tumor, CSF, and white matter) and all spatial tokens within a slice. As illustrated in Fig.~\ref{fig:rep_structure}(a), BrainDINO produced spatially selective activation patterns that closely corresponded to anatomical compartments---tumor-centered references yielded segmentation-like highlights across enhancing lesions, while CSF and white matter references selectively activated their respective tissue types---all without any task-specific supervision. This suggests that jointly optimizing for global semantic consistency and local structural fidelity during pretraining produces a spatially precise and anatomically organized feature geometry.

To assess cross-model representational consistency and how feature geometry evolves across network depth, we performed layer-wise centered kernel alignment (CKA)~\cite{kornblith2019similarity} on both normative (IXI) and pathological (BraTS) cohorts (Fig.~\ref{fig:rep_structure}(b, right)). Among all comparators, BrainDINO remained most closely aligned with DINOv3, with CKA increasing across network depth in both cohorts. This alignment reflects the shared representational geometry induced by the DINO training objective. On the normative IXI cohort, similarity was high throughout and reached 0.953 at the late layer. On the pathological BraTS cohort, CKA with DINOv3 was systematically lower at every matched depth (0.523 vs. 0.811 at the early layer and 0.837 vs. 0.953 at the late layer), indicating reduced alignment with natural-image representations under pathological distribution shift. In contrast, MRI-specific SSL baselines exhibited substantially lower and less stable CKA similarity to BrainDINO across all layers and both cohorts, indicating that these models occupy a different region of representational space.

\section{Discussion}

We developed a large-scale slice-wise self-supervised foundation model for brain MRI, pretrained on 6.6 million unlabeled axial slices collected from 20 heterogeneous datasets spanning normative populations, neurodevelopmental cohorts, structural brain tumor imaging, and diverse clinical acquisition settings. Without any task-specific supervision during pretraining, the learned representation consistently improves performance across a broad spectrum of downstream applications, including tumor segmentation, neurodevelopmental and neurodegenerative classification, neuroanatomical trajectory modeling, survival risk stratification, mutation detection, and MRI sequence classification. These improvements are observed across tasks with varying levels of clinical complexity and dataset scale, with particularly pronounced gains under limited-label regimes. Because brain MRI exhibits statistical and structural properties distinct from natural images, these improvements over a natural-image-pretrained backbone on the majority of tasks suggest that large-scale brain-specific self-supervised learning can yield a generalizable representation capable of supporting diverse neuroimaging tasks with minimal downstream adaptation. The principal exception is MRI sequence classification, an acquisition-level task on which both slice-based ViTs---brain-specific and natural-image-pretrained---saturate and perform comparably, indicating that the benefit of brain-specific pretraining is largest for tasks requiring fine anatomical or pathological discrimination rather than global contrast cues.

A distinguishing characteristic of our framework is its parameter-efficient adaptation strategy. Across classification and regression tasks, only approximately 0.6\% of total model parameters are updated during downstream training, while over 99\% of the pretrained encoder remains frozen. For dense segmentation, the backbone likewise remains frozen (77.4\% of total parameters), with only the task-specific decoder and adapter modules (22.6\%) optimized. This design aligns with the intended philosophy of foundation models~\cite{bommasani2021opportunities}, wherein a pretrained representation functions as a stable, reusable backbone rather than being re-specialized per task. From a clinical deployment perspective, parameter-efficient adaptation reduces computational demands, shortens fine-tuning time, and mitigates representation drift when transferring to new institutions or limited-label cohorts. While several prior brain MRI foundation approaches, including
BrainIAC~\cite{tak2026generalizable}, adopt full-network fine-tuning,
our results indicate that a well-structured brain-specific representation
can support both linear probing and dense prediction with minimal
task-specific modification.
To directly quantify this trade-off, we compared frozen-backbone and full 
fine-tuning adaptation on three representative tasks spanning distinct 
supervision types and parameter budgets (Supplementary Table~6). The two 
strategies yielded comparable performance across the majority of conditions, 
with the frozen encoder remaining competitive or superior under limited 
supervision and matching full fine-tuning on dense segmentation even at 
full data availability. These results indicate that representation quality, 
rather than end-to-end gradient flow, is the primary driver of downstream 
performance, and support frozen-backbone adaptation as a default and 
practically viable strategy for clinical deployment.

Beyond parameter efficiency, BrainDINO demonstrates consistent data efficiency across diverse downstream tasks. Performance advantages persist even under extremely limited supervision (e.g., 10\% and 20\% labeled data), with relative gains particularly evident in clinically challenging, data-scarce settings. Improvements span linear classification, regression, and dense segmentation, indicating that the pretrained representation retains sufficient structural information to support task adaptation in low-data regimes, which is a practical advantage frequently needed in clinical neuroimaging. This behavior is further supported by representation-level analyses. Non-parametric kNN probing on frozen features indicates stronger empirical class separability prior to any task-specific adaptation, while reference-based similarity maps and layer-wise CKA reveal that the learned representation is both spatially structured and hierarchically organized. 

These findings suggest that pretraining on
heterogeneous brain MRI data at scale induces invariances that
suppress acquisition-driven distributional variation while preserving
clinically meaningful structure. The stability observed under
test-time perturbations simulating scanner and protocol differences
provides additional evidence for this property. These invariances in
turn appear to support a shared anatomical encoding from which
segmentation, classification, regression, and survival modeling can be
realized as different projections without task-specific encoder
adaptation. A formal characterization of this representational
geometry remains an important direction for future investigation.

Several limitations should be acknowledged. First, the slice-wise formulation may limit inter-slice contextual continuity; qualitative inspection suggests that segmentation predictions tend to be conservative on slices with less prominent pathology, and future work may explore cross-slice aggregation or fully 3D pretraining. Second, certain applications, such as autism spectrum disorder 
classification, Parkinson's disease classification, and post-stroke 
temporal estimation, remain challenging, likely reflecting subtle or 
diffuse imaging signatures. For Parkinson's disease, whose pathology is 
predominantly dopaminergic, prior work has largely relied on 
diffusion-based white-matter and microstructural 
features~\cite{koirala2019alterations} rather than T1 macrostructure, 
yet BrainDINO retained a consistent advantage using only frozen T1 
representations. These cases may benefit from domain-specific 
architectures or multimodal integration. Third, we intentionally preserved a frozen backbone to emphasize representation stability; systematic evaluation of full-network fine-tuning may yield additional gains at the expense of computational cost and transfer stability. 

Looking ahead, several directions remain open for future work. At the 
downstream level, deformable registration~\cite{balakrishnan2019voxelmorph} 
was excluded from our evaluation because of its limited performance, and 
warrants further investigation. At the pretraining level, our corpus is 
restricted to structural MRI sequences; broadening it to additional 
modalities such as functional MRI and diffusion-weighted imaging, or 
coupling our representation with existing self-supervised foundation 
models of complementary modalities, could extend its scope considerably. 
Beyond brain imaging, these include functional MRI~\cite{wei2026fmri} and 
electrophysiological signals such as EEG and MEG~\cite{xiao2026brainomni}, 
as well as peripheral physiological signals that reflect central autonomic 
function, such as ECG~\cite{monachino2025self}. Integrating structural MRI 
with such complementary signals points toward unified, multimodal 
foundation models that jointly capture brain structure, function, and 
neurophysiological state---an integrative direction increasingly central 
to data-driven precision medicine.

\section{Methods}
\subsection{Pretraining Data}

\subsubsection{Dataset Collection}

We construct a large-scale pretraining corpus by aggregating publicly available brain MRI datasets
that collectively capture broad variation in population cohorts, disease entities, and imaging settings.
The corpus integrates normative population studies, including ICBM~\cite{mazziotta2001probabilistic},
ID1000 and PIOP1\&2~\cite{snoek2021amsterdam}; lifespan and neurodevelopmental cohorts such as
LONG579~\cite{wang2022longitudinal}, ABIDE~\cite{di2014autism}, and PPMI~\cite{marek2011parkinson}; structural brain tumor
datasets, including TCGA-GBM~\cite{scarpace2016cancer}, CPTAC-GBM~\cite{cptac2018gbm}, GLIS-RT~\cite{shusharina2021glis},
REMBRANDT~\cite{scarpace2019rembrandt}, IvyGAP~\cite{puchalski2018anatomic}, BraTS2025-GLI-PRE~\cite{de20242024},
BraTS2023-PED~\cite{kazerooni2024brain}, BraTS2023-MEN~\cite{labella2024multi}, BraTS2023-SSA~\cite{adewole2023brain},
MEN-SEG-CLASS~\cite{vassantachart2023meningioma}, and ReMIND~\cite{juvekar2024remind}; as well as clinical acquisition and workflow datasets
such as fastMRI-Brain~\cite{zbontar2018fastmri}, ACRIN-DSC-MR-Brain~\cite{kinahan2019acrin}, and ACRIN-FMISO-Brain~\cite{gerstner2012acrin}.
In addition, multiple multi-center clinical trial cohorts are included to further enhance diversity
in scanner hardware and acquisition protocols.

In total, the pretraining corpus comprises 20 heterogeneous datasets and about 6.6 million 2D MRI slices extracted from over 51k 3D volumes. As summarized in Fig.~\ref{fig:overview}, the slice-scale distribution spans several orders of magnitude across datasets, while collectively covering a wide range of subject demographics, imaging protocols, and clinical contexts. The resulting corpus encompasses neurodevelopmental, neurodegenerative, and oncological conditions, as well as variations driven by acquisition and reconstruction differences, providing a rich substrate for learning anatomy-aware and disease-agnostic representations.

All datasets are used exclusively in an unlabeled manner during pretraining. For datasets that also appear in downstream evaluation, pretraining and evaluation are conducted under strict subject-level separation: no subject in the downstream test set was seen during pretraining. Detailed overlap descriptions are provided in each downstream task section.

\subsubsection{Slice-wise Sampling Strategy}
 
We extracted 2D axial slices from 3D brain MRI volumes as the basic training units for self-supervised pretraining. Each volume was processed through a sequential pipeline prior to slice extraction. First, skull stripping was performed using HD-BET~\cite{isensee2019automated} to remove non-brain tissues, ensuring that sampled slices primarily captured intracranial anatomy. This step standardized the foreground definition across datasets with heterogeneous acquisition protocols and fields of view, and was applied uniformly without using any task-specific information. Each skull-stripped volume was then cropped to the minimal bounding box enclosing non-zero voxels and resized to a fixed spatial resolution of $256 \times 256 \times 256$.
 
Intensity values were subsequently standardized at the volume level using percentile-based normalization: voxel intensities were clipped to the $0.5$th--$99.5$th percentile range to suppress extreme outliers and scanner-specific intensity spikes, followed by $z$-score normalization using the mean and standard deviation of non-zero voxels within the volume. This procedure reduced inter-dataset intensity variability while preserving relative anatomical contrast. After normalization, 128 axial slices were sampled from the 256 available slices along the superior-to-inferior axis. To avoid over-representing slices near the superior and inferior extremes, which often contained limited anatomical or clinical information, slices were selected according to a one-dimensional normal distribution centered at the midpoint of the axis ($\mu = 128$, $\sigma = 50$). This strategy biased sampling toward central brain regions, where clinically relevant structures and pathological patterns were more frequently observed, while still preserving coverage across the full cranio-caudal range.
 
Unless otherwise specified, this preprocessing and sampling pipeline was applied consistently across all pretraining datasets, enabling large-scale and dataset-agnostic representation learning while maintaining a balance between anatomical coverage and information density.

\subsection{Self-Supervised Pretraining Framework}
We adopted a teacher-student self-distillation framework inspired by DINOv3 for slice-wise self-supervised pretraining on brain MRI (Fig.~\ref{fig:overview}(b)). The model was trained to produce consistent representations across multiple augmented views of the same 2D MRI slice without relying on manual annotations.

The student and teacher networks shared an identical Vision Transformer backbone (ViT-B/16; patch size $16$) with rotary positional embeddings and $4$ learned storage tokens. Each input slice was replicated to three channels to conform to the RGB input convention of DINOv3 and fed to both networks. From each slice, we generated a multi-crop set consisting of two global views and eight local views. Global crops were sampled with scale range $[0.32, 1.0]$ and resized to $256\times256$, while local crops were sampled with scale range $[0.05, 0.32]$ and resized to $112\times112$. To encourage invariance to acquisition noise and minor geometric perturbations in grayscale MRI, we applied random horizontal flips and Gaussian blurring. The teacher network received only the two global views, while the student network was trained on all views (two global plus eight local), encouraging view-invariant representations under both large-field and localized perturbations.

The teacher parameters were updated as an exponential moving average (EMA) of the student parameters, which stabilized optimization and mitigated representation collapse during training. At each iteration, teacher weights $\theta_t$ were updated from student weights $\theta_s$ using momentum $m$ fixed to $0.994$ throughout training. Each network produced both a global class (CLS) token and patch-level token embeddings, which were passed through lightweight projection heads prior to computing self-distillation losses. Specifically, we used separate DINO and iBOT heads (three-layer MLP; hidden dimension $2048$; bottleneck dimension $256$) with a prototype dimension of $K=65536$ for both objectives.

Unlike the full DINOv3 formulation, we did not incorporate Gram matrix regularization. In preliminary experiments on heterogeneous brain MRI data, Gram-based feature regularization did not yield consistent improvements and occasionally led to reduced downstream performance. To preserve representation stability and generality across datasets, we retained only the core self-distillation objectives.

Pretraining jointly optimized global and local consistency terms. Global semantic alignment was enforced through a DINO-style cross-entropy loss applied to the CLS tokens, where the student output distribution was matched to a temperature-sharpened and centered teacher distribution:

\begin{equation}
\mathcal{L}_{\mathrm{DINO}} = - \sum_{k} q_k^{(t)} \log p_k^{(s)},
\end{equation}
where $q^{(t)}$ denotes the teacher probability distribution and $p^{(s)}$ denotes the student distribution.

In parallel, local structural consistency was encouraged through an iBOT-style~\cite{zhou2021ibot} masked prediction objective applied to patch tokens. For each global crop, a subset of patch tokens was randomly masked with probability $0.5$, using a mask ratio uniformly sampled from $[0.1, 0.5]$. The student predicted the teacher's centered targets only at masked locations:
\begin{equation}
\mathcal{L}_{\mathrm{iBOT}} = - \sum_{i \in \mathcal{M}} \sum_{k} q_{ik}^{(t)} \log p_{ik}^{(s)},
\end{equation}
where $\mathcal{M}$ denotes the set of masked patch locations. The overall training objective combined these two components:
\begin{equation}
\mathcal{L} = \mathcal{L}_{\mathrm{DINO}} + \lambda \mathcal{L}_{\mathrm{iBOT}},
\end{equation}
with $\lambda=1$ in all experiments. To further regularize the representation geometry, we additionally applied a KoLeo diversity loss on the student CLS features with weight $0.1$, encouraging non-collapsed and well-spread embeddings.

Optimization was performed using AdamW with gradient clipping (max norm $3.0$) and layer-wise learning-rate decay ($0.9$). We trained for $500,000$ iterations. Training used 4 A100 GPUs and a per-GPU batch size of $256$. The learning rate was warmed up for $10,000$ iterations followed by cosine decay, using a peak base learning rate of $10^{-4}$. Weight decay was held constant at $0.04$ throughout training. 

To improve representation for fine anatomical structures, we performed an additional high-resolution stage after the initial pretraining. Each $256 \times 256$ slice was upsampled via interpolation to $1024 \times 1024$ before cropping. This stage continued self-supervised learning under a multi-resolution crop setting. Specifically, five global--local crop pairs were jointly sampled during training, with global crop sizes of $\{512, 768, 768, 768, 768\}$ and corresponding local crop sizes of $\{112, 112, 168, 224, 336\}$. Sampling ratios across these crop pairs were set to $\{0.30, 0.30, 0.30, 0.05, 0.05\}$. This design allowed the model to adapt to varying high-resolution spatial structures. The per-GPU batch size was $56$ and training continued for $10,000$ iterations.

Pretraining was conducted exclusively on 2D axial slices rather than full 3D volumes. This slice-wise design enabled efficient scaling across heterogeneous datasets with varying voxel spacings and acquisition protocols, while avoiding assumptions about volumetric alignment across cohorts. By treating slices as independent training units, the model was exposed to large-scale anatomical variability and learned anatomy-aware yet disease-agnostic representations that remained transferable across downstream tasks.

\subsection{Downstream Task Finetuning}

For all downstream tasks, we fine-tuned lightweight task-specific heads on top of a frozen BrainDINO backbone (Fig.~\ref{fig:overview}(c), middle). Unless otherwise specified, all tasks shared a unified slice-wise feature extraction pipeline. Each 3D MRI volume was resampled to a spatial resolution of $128 \times 256 \times 256$, uniformly sampled into 128 axial slices, processed independently by the pretrained 2D ViT encoder, and aggregated into subject-level representations via mean pooling of slice-wise CLS tokens. This design isolated the contribution of the pretrained representation and ensured fair comparison across different pretraining strategies.

Across tasks, the head shared a common lightweight architecture consisting of LayerNorm, a linear projection with GELU activation, and a final linear layer, producing $K$-class logits for classification and a single continuous value for regression. Task-specific loss functions are defined in their respective subsections below. Unless otherwise noted, training used the Adam optimizer with a learning rate of $1\times10^{-4}$ and weight decay of $1\times10^{-5}$.

\subsubsection{Tumor Segmentation}

Tumor segmentation was evaluated using a frozen-encoder paradigm within the nnU-Net framework across four brain tumor benchmarks: BraTS2021 (glioma), BraTS2023-Mets (brain metastases), BraTS2023-MEN (meningioma), and BraTS2024-GoAT (multi-tumor generalizability). All four datasets provide four MRI modalities (T1, T1c, T2, FLAIR) and are evaluated on three composite tumor regions, namely whole tumor (WT), tumor core (TC), and enhancing tumor (ET), assembled from dataset-specific raw annotations (Supplementary Table 5).

For BrainDINO and the DINOv3 baseline, a 2D slice-based pipeline was used with nnU-Net configuration \texttt{2d} at $256 \times 256$ resolution. The four MRI modalities were handled jointly via a learnable $1{\times}1$ convolution mapping 4 input channels to 3, initialized as an equal-weight average. The frozen ViT-B encoder was paired with a Primus decoder that extracted intermediate features from transformer layers $\{2, 5, 8, 11\}$. Of the 110.7\,M total parameters, 77.4\% belonged to the frozen encoder and 22.6\% to the trainable decoder and input fusion layer. Three 3D baselines operated on volumetric patches with their respective frozen encoders and compatible decoders: BrainMVP and BM-MAE used $128^3$ patches, while BrainIAC used its native $96^3$ patches; pretrained single-channel patch embeddings were adapted to 4-channel input by repeating and rescaling weights.

All experiments followed a data-ratio sweep over $\{0.1, 0.2, 0.4, 0.6, 0.8, 1.0\}$ of the training split using nnU-Net fold~0, with validation splits held constant. Training minimized a Dice~+~cross-entropy loss using AdamW. BrainDINO trained for 100 epochs with gradient clipping (max norm 12) and mixed-precision training. Segmentation accuracy was reported as mean Dice $\pm$ standard deviation per region. Each training epoch comprised 250 training and 50 validation 
iterations across all methods, with deep supervision disabled 
and standard nnU-Net augmentations applied throughout. The 3D 
baselines (BrainIAC, BrainMVP, BM-MAE) followed the nnU-Net 
default PolyLR schedule parameterized over 1000 epochs while 
running for 100 epochs in practice. For BraTS2023-MEN, 200 
cases used during pretraining were drawn exclusively from 
the downstream training split (fold~0), with no overlap in 
the held-out test set.

\subsubsection{Neurodevelopmental and Neurodegenerative Classification}
Neurodevelopmental and neurodegenerative classification was formulated as a 
supervised task using structural T1-weighted brain MRI to evaluate whether the 
learned representation captured disease-related neuroanatomical patterns. 
Experiments were conducted on four cohorts. For ABIDE, binary ASD classification 
contrasted individuals with ASD and neurotypical controls, with 492 controls and 
464 ASD subjects for training (956 total) and 123 controls and 117 ASD subjects 
for testing (240 total); among these, 853 subjects used in pretraining were drawn 
exclusively from the downstream training cohort, with no overlap with the test 
set. For ADNI, three-class Alzheimer's disease staging spanned cognitively normal
(CN), mild cognitive impairment (MCI), and Alzheimer's disease (AD); the
cohort comprises 2{,}182 T1-weighted scans from 382 subjects (CN/MCI/AD =
748/981/453 scans across longitudinal visits sc, m06--m36). Because each
subject contributes multiple visits with an essentially constant diagnostic
label, ADNI is evaluated under a strict subject-level protocol: a single
scan-level train/test split would place different visits of the same
subject in both partitions and inflate the metric. We therefore use
subject-disjoint cross-validation in which all scans of a subject are kept
together, so that no subject appears in both the training and test
sets. Reported macro-AUC at each labeled-data
ratio is the mean across the test splits (95\% confidence intervals are
Student-$t$ intervals over the splits; significance against
BrainDINO is a paired test across splits). In addition to the three-class
task, two clinically motivated binary contrasts were computed on the same
splits: CN-vs-AD (the standard ADNI binary task, MCI subjects excluded) and
CN-vs-MCI (the more challenging early-detection task of distinguishing cognitively normal individuals from those with subtle prodromal cognitive impairment).
Age-stratified analysis at $100\%$ supervision was computed from the
\emph{pooled subject-disjoint test predictions}:
each scan's predicted probability is the mean across the splits in which
it was held out, and macro-AUC is then re-computed independently within
each age stratum (55--64, 65--74, 75--84, 85+). Per-stratum estimates
therefore reflect held-out predictions only and are not measured on
training data; scans not selected into any
test split are excluded from age-stratified analysis. For OASIS, dementia was classified from 
clinical dementia rating into non-demented and demented, with 106 and 89 subjects 
for training (195 total) and 29 and 11 subjects for testing (40 total). For PPMI, 
binary PD classification contrasted patients with PD against neurologically 
healthy controls using baseline T1-weighted MRI, with 1158 PD and 283 control 
subjects; because PD pathology is predominantly dopaminergic rather than 
macrostructural, structural MRI carries limited discriminative signal for this 
task, and the pronounced class imbalance (approximately 4:1) was addressed by 
applying synthetic minority oversampling to the training features while the test 
set was left unmodified. For all cohorts, no subject in the held-out test set was 
seen during pretraining.

The classification head was trained by minimizing the cross-entropy loss,
\begin{equation}
\label{eq:ce}
\mathcal{L}_{\mathrm{CE}} = -\log \frac{\exp(z_y)}{\sum_{k=1}^{K} \exp(z_k)},
\end{equation}
where $\mathbf{z} = (z_1,\ldots,z_K)$ denotes the output logits and $y$ the 
ground-truth label, with $K=2$ for the binary cohorts (ABIDE, OASIS, PPMI) and 
$K=3$ for ADNI. For PPMI, subject-level representations were aggregated from 
patch-token features over a fixed central axial range corresponding to subcortical 
structures most affected in PD. Performance was evaluated using ROC-AUC, reported 
as standard ROC-AUC for the binary cohorts and macro-averaged AUC for the 
multi-class ADNI setting, across training data ratios from $10\%$ to $100\%$.
\subsubsection{Neuroanatomical Trajectory Modeling}

Neuroanatomical trajectory modeling tasks were conducted to evaluate whether the learned representation captured temporally structured anatomical patterns in brain MRI that reflect physiological aging and pathological evolution following injury.

Brain age prediction was formulated as a continuous regression task to assess sensitivity to aging-related neuroanatomical variation. Experiments were conducted using MRI data from the IXI, LONG579, and Pixar datasets, spanning a broad age range. The regression head was trained to predict chronological age by minimizing the mean squared error (MSE):
\begin{equation}
\mathcal{L}_{\mathrm{MSE}} = (\hat{y} - y)^2.
\end{equation}
Performance was evaluated using mean absolute error (MAE),
\begin{equation}
\mathrm{MAE} = \frac{1}{N} \sum_{i=1}^{N} \left| \hat{y}_i - y_i \right|,
\end{equation}
providing an interpretable measure of prediction error in years.

We further evaluated post-stroke temporal modeling on the ATLAS dataset by predicting Days Post Stroke (DPS) as a continuous variable. To stabilize optimization under skewed label distributions, the regression target was transformed during training as
\begin{equation}
\tilde{y} = \log(1 + y),
\end{equation}
with inverse transformation applied at inference,
\begin{equation}
\hat{y} = \exp(\hat{y}_{\log}) - 1.
\end{equation}
A mixed regression objective was adopted to balance sensitivity and robustness,
\begin{equation}
\mathcal{L}_{\text{ATLAS}} =
\alpha (\hat{y}_{\log} - \tilde{y})^2
+
(1-\alpha)
\left(
0.5\, |\hat{y}_{\log} - \tilde{y}|
+
0.5\, \mathrm{Huber}_\delta(\hat{y}_{\log} - \tilde{y})
\right),
\end{equation}
where $\alpha = 0.2$ and $\delta = 1.0$. AdamW was used for training with a learning rate of $10^{-3}$ and weight decay of $10^{-4}$, and performance was evaluated using MAE, RMSE, and $R^2$.

\subsubsection{Survival Prediction}

Survival prediction was evaluated on the UPENN-GBM cohort to assess whether the learned representation supported prognostic modeling under both classification-based and time-to-event formulations.

For binary survival prediction, patients were stratified using a one-year threshold into low-risk ($>365$ days) and high-risk ($\leq 365$ days) groups. The training cohort included 270 low-risk and 212 high-risk patients (482 total), and the testing cohort included 68 low-risk and 53 high-risk patients (121 total). The classification head was trained with $K=2$ by minimizing the cross-entropy loss defined above. Performance was evaluated using ROC-AUC.

Time-to-event survival analysis was further conducted using a Cox proportional hazards formulation. For multi-modal survival modeling, only patients with complete four-modality MRI (T1, T1CE, T2, and FLAIR) and available clinical follow-up were included, resulting in 414 training and 106 testing subjects after filtering from the original split. The training cohort comprised 233 censored and 181 deceased patients, and the testing cohort included 56 censored and 50 deceased patients. In this formulation, event $= 1$ denoted deceased status and event $= 0$ denoted censored (lost to follow-up), and patients censored within one year were excluded from analysis. Each modality was processed independently through the frozen BrainDINO encoder using the same slice-wise pipeline. Modality-specific representations were fused via late fusion by averaging modality-wise features, and a regression head output a continuous risk score $h_i$ for each patient. The hazard function was defined as

\begin{equation}
\lambda(t \mid X_i) = \lambda_0(t)\exp(h_i),
\end{equation}
where $\lambda_0(t)$ denotes the baseline hazard. Optimization minimized the Cox partial likelihood,
\begin{equation}
\mathcal{L}_{\mathrm{Cox}} =
-\frac{1}{D}
\sum_{i:E_i=1}
\left[
h_i -
\log
\sum_{j:T_j \geq T_i}
\exp(h_j)
\right],
\end{equation}
using Adam with learning rate $3\times10^{-4}$ and weight decay $1\times10^{-3}$. Performance was evaluated using the concordance index (C-index), measuring agreement between predicted risk rankings and observed survival times.

\subsubsection{Mutation Detection}

Isocitrate dehydrogenase (IDH) mutation status prediction was formulated as a binary classification task using the UCSF-PDGM dataset. After filtering for cases with both T1CE and FLAIR modalities available, the cohort comprised 360 training and 92 testing subjects. The training set included 288 IDH wildtype and 72 non-wildtype cases, and the testing set included 73 wildtype and 19 non-wildtype cases. The non-wildtype group included mutated, unknown, and other IDH subtypes as defined in the dataset construction.

Two MRI modalities (T1CE and T2-FLAIR) were processed independently through the frozen BrainDINO encoder using the same slice-wise pipeline, and modality-level representations were fused via feature averaging. The classification head with $K=2$ and dropout regularization was trained by minimizing the cross-entropy loss. Performance was evaluated using ROC-AUC.

\subsubsection{MRI Sequence Classification}

MRI sequence classification was performed to assess whether the pretrained representation encoded acquisition-level semantics. This task was formulated as a four-way classification problem ($K=4$) to distinguish between T1CE, T1, FLAIR, and T2 sequences using the BraTS-2023 dataset.

The combined cohort included 3853 training and 971 testing volumes (4824 total). The testing set included 249 T1CE, 239 T1, 239 FLAIR, and 244 T2 volumes. The dataset was aggregated from three BraTS-2023 sub-cohorts: BraTS2023-MEN (4188 total volumes, 86.8\%), BraTS2023-SSA (240 volumes, 5.0\%), and BraTS2023-PED (396 volumes, 8.2\%), preserving their predefined training and testing splits. Among these, 200 volumes from BraTS2023-MEN, 120 from BraTS2023-SSA, and 100 from BraTS2023-PED were seen during pretraining; these overlapping cases were used only in the downstream training set and were not included in the test set.

The classification head was trained using Adam with a learning rate of $1\times10^{-4}$ and weight decay of $1\times10^{-5}$. Performance was assessed using accuracy and macro-averaged ROC-AUC.
\subsection{Frozen-Backbone versus Full Fine-Tuning Comparison}

To assess whether freezing the pretrained encoder limited downstream performance, we conducted an additional comparison between frozen-backbone adaptation and full fine-tuning using BrainDINO on three representative tasks: ADNI three-class classification, brain age regression, and BraTS2021 tumor segmentation. These tasks were selected to span classification, continuous regression, and dense prediction. Experiments were performed under 10\% and 100\% labeled-data availability using the same train/test splits, preprocessing, task heads or decoders, loss functions, and evaluation metrics as the corresponding frozen-backbone experiments. In the frozen setting, the pretrained encoder was fixed and only the task-specific head or decoder was optimized; in the full fine-tuning setting, the encoder and task-specific modules were optimized jointly. Statistical comparisons followed the same task-specific procedures used in the main downstream evaluation.

\subsection{Robustness Analysis}

We evaluated test-time robustness across three representative downstream task types: tumor segmentation (BraTS2021), neurodegenerative classification (ADNI), and brain age regression. These tasks span dense spatial prediction, categorical inference, and continuous biomarker estimation.

All four brain-MRI-pretrained models (BrainDINO, BrainIAC, BM-MAE, and BrainMVP) were assessed at full-data training ratio (1.0); the natural-image-pretrained DINOv3 was not included in the perturbation analysis. Segmentation performance was measured using Dice for WT, TC, and ET; classification was evaluated using macro-AUC; regression was evaluated using MAE.

Perturbations were applied to raw MRI volumes prior to preprocessing and inference, without additional training or test-time adaptation. Only one perturbation type was applied at a time. We examined contrast shift via gamma correction ($\gamma \in \{0.5,0.8,1.0,1.2,1.5,2.0\}$), Gibbs artifact via k-space truncation ($\alpha \in \{0.0,0.1,0.2,0.3,0.4\}$), and bias field via smooth multiplicative RF inhomogeneity ($s \in \{0.0,0.1,0.2,0.3,0.4\}$), where the zero or unity value corresponds to the clean condition.

\subsection{Representation-Level Analysis}

\paragraph{kNN probing}
Frozen CLS-token features were extracted using the slice-wise pipeline described above, where each subject was represented by the mean of 128 uniformly sampled slice-level CLS tokens. All features were L2-normalized, and kNN classification was performed using cosine similarity. The training split was used as the memory bank and the test split as queries under a fixed protocol. Predictions were obtained via majority voting, with ties broken by the summed cosine similarity within tied classes. We swept $k \in \{1, 3, 5, 10, 20, 50\}$ to assess neighborhood stability. No PCA or whitening was applied.

\paragraph{Reference similarity maps}
Spatial similarity maps were computed from patch-token embeddings. Given a reference token $f_{\mathrm{ref}}$ and a spatial token $f_{i,j}$, similarity was defined as cosine similarity between their L2-normalized features:
\begin{equation}
s(i,j) = \frac{f_{\mathrm{ref}} \cdot f_{i,j}}{\|f_{\mathrm{ref}}\| \, \|f_{i,j}\|}.
\end{equation}
Three anatomically defined reference locations (tumor, CSF, and white matter) were selected on a fixed slice, and identical pixel coordinates were used across all encoders. For 2D encoders (BrainDINO, DINOv3), similarity was computed 
over the native 2D patch-token grid; for 3D encoders (BM-MAE, 
BrainMVP, BrainIAC), the corresponding axial token plane was 
identified by mapping the slice index to the nearest depth 
position in the 3D token grid. Similarity maps were bilinearly upsampled to slice resolution and visualized using a shared color scale without per-encoder rescaling.

\paragraph{CKA analysis}
We used linear centered kernel alignment (CKA)~\cite{kornblith2019similarity} 
to compare subject-level representations between BrainDINO and 
four alternative pretrained backbones (DINOv3, BM-MAE, BrainMVP, 
BrainIAC) at three representative depths corresponding to early 
(block 2), mid (block 6), and late (block 11) layers. For each 
model pair, Gram matrices $K = XX^\top$ and $L = YY^\top$ were 
computed from subject-level feature matrices $X$ and $Y$ and 
centered with $H = I - \mathbf{1}\mathbf{1}^\top/n$. For the 
2D encoders (BrainDINO, DINOv3), 128 axial slices per volume 
were resized to $224 \times 224$ and replicated to three 
channels; slice-level CLS tokens were then averaged to obtain 
one subject-level feature vector per layer. For BM-MAE and 
BrainIAC, CLS-token features at the corresponding blocks were 
used directly; for BrainMVP, globally average-pooled spatial 
features from early, mid, and late stages were used. CKA was 
computed over a fixed subset of 1{,}000 volumes per dataset 
without shuffling: the IXI subset comprised 502 T1 and 498 T2 
volumes from 502 unique subjects, and the BraTS2021 subset 
comprised 250 subjects with four modalities each.

\subsection{Statistics and Reproducibility}
Statistical comparisons across downstream tasks were tailored 
to the inference target. For classification tasks (ABIDE, ADNI, 
OASIS, IDH mutation, MRI sequence, binary survival) and dense 
segmentation (BraTS benchmarks), pairwise differences against 
BrainDINO at matched labeled-data ratios were assessed via 
paired bootstrap resampling of the test set, with 
significance evaluated as the proportion of resampled differences 
crossing zero. For continuous regression (brain age, post-stroke 
DPS), pairwise differences in MAE were assessed via paired 
$t$-tests on per-subject absolute errors. Survival risk 
stratification on UPENN-GBM was evaluated using two-sided 
log-rank tests on Kaplan--Meier curves. Frozen-feature kNN 
classification accuracy comparisons used the exact McNemar test 
(two-sided binomial test on discordant pairs) with Holm 
correction across neighborhood sizes; corresponding accuracy 
confidence intervals were obtained from stratified 
test-set bootstrap resampling with 2{,}000 iterations 
(seed 42, 2.5th/97.5th percentiles). Age-stratified analyses 
applied Holm-adjusted pairwise tests within each age bin: 
pairwise bootstrap for AD diagnosis on ADNI and paired 
sign-flip tests for brain age estimation. Unless otherwise 
specified, the significance threshold was $\alpha = 0.05$ and 
all tests were two-sided. Significance levels are annotated 
in figures and Supplementary Tables using a tiered notation 
($^{*}p < 0.05$, $^{**}p < 0.01$, $^{***}p < 0.001$); 
in-text comparisons are stated as $p < 0.05$ for narrative 
clarity. Uncertainty is reported as mean $\pm$ standard 
deviation for segmentation Dice and regression MAE, and as 
point estimates with 95\% bootstrap confidence intervals 
(in square brackets) for AUC-based classification, mutation, 
and survival metrics.

\section{Data availability}
This study used exclusively publicly available brain MRI datasets for both pretraining and downstream evaluation, as detailed in Supplementary Table~2 and Methods Section~4.1. Publicly available datasets include OASIS-1/2/3 (\url{https://sites.wustl.edu/oasisbrains/}), IXI (\url{https://brain-development.org/ixi-dataset/}), ABIDE (\url{http://fcon_1000.projects.nitrc.org/indi/abide/}), ADNI (\url{https://adni.loni.usc.edu/}), and BraTS2021/2023/2024 (\url{https://www.synapse.org/}), among others as listed in Supplementary Table~2. Some datasets require registration and data use agreements prior to access. Source data for all figures and tables are provided in the Supplementary Materials.

\section*{Code availability}
The downstream task code for BrainDINO is available at 
\url{https://github.com/mclwu22/BrainDINO}. The pretraining 
code and pretrained model weights will be released upon 
acceptance, and are available to Editors and reviewers upon 
request during the peer-review process.

\section*{Funding}
This research is supported in part by the National Institutes of Health 
under Award Numbers R01EB032680, R01CA272991, and U54CA274513.

\section*{Competing interests}
The authors declare no competing financial or non-financial 
competing interests.

\section*{Ethics statement}
This study used exclusively de-identified, publicly available 
brain MRI datasets that had been previously approved by the 
respective institutional review boards of the contributing 
institutions. No additional ethical approval was required for 
the present analysis.




\bibliography{sn-bibliograph}

@article{menze2014multimodal,
  title={The multimodal brain tumor image segmentation benchmark (BRATS)},
  author={Menze, Bjoern H and Jakab, Andras and Bauer, Stefan and Kalpathy-Cramer, Jayashree and Farahani, Keyvan and Kirby, Justin and Burren, Yuliya and Porz, Nicole and Slotboom, Johannes and Wiest, Roland and others},
  journal={IEEE transactions on medical imaging},
  volume={34},
  number={10},
  pages={1993--2024},
  year={2014},
  publisher={IEEE}
}

@article{bakas2017advancing,
  title={Advancing the cancer genome atlas glioma MRI collections with expert segmentation labels and radiomic features},
  author={Bakas, Spyridon and Akbari, Hamed and Sotiras, Aristeidis and Bilello, Michel and Rozycki, Martin and Kirby, Justin S and Freymann, John B and Farahani, Keyvan and Davatzikos, Christos},
  journal={Scientific data},
  volume={4},
  number={1},
  pages={170117},
  year={2017},
  publisher={Nature Publishing Group}
}

@article{isensee2018nnu,
  title={nnu-net: Self-adapting framework for u-net-based medical image segmentation},
  author={Isensee, Fabian and Petersen, Jens and Klein, Andre and Zimmerer, David and Jaeger, Paul F and Kohl, Simon and Wasserthal, Jakob and Koehler, Gregor and Norajitra, Tobias and Wirkert, Sebastian and others},
  journal={arXiv preprint arXiv:1809.10486},
  year={2018}
}

@inproceedings{suk2013deep,
  title={Deep learning-based feature representation for AD/MCI classification},
  author={Suk, Heung-Il and Shen, Dinggang},
  booktitle={International conference on medical image computing and computer-assisted intervention},
  pages={583--590},
  year={2013},
  organization={Springer}
}

@article{ebrahimi2021convolutional,
  title={Convolutional neural networks for Alzheimer's disease detection on MRI images},
  author={Ebrahimi, Amir and Luo, Suhuai and Disease Neuroimaging Initiative, for the Alzheimer's},
  journal={Journal of Medical Imaging},
  volume={8},
  number={2},
  pages={024503--024503},
  year={2021},
  publisher={Society of Photo-Optical Instrumentation Engineers}
}

@article{dardouri2025efficient,
  title={An efficient method for early Alzheimer's disease detection based on MRI images using deep convolutional neural networks},
  author={Dardouri, Samia},
  journal={Frontiers in Artificial Intelligence},
  volume={8},
  pages={1563016},
  year={2025},
  publisher={Frontiers Media SA}
}

@article{cole2017predicting,
  title={Predicting brain age with deep learning from raw imaging data results in a reliable and heritable biomarker},
  author={Cole, James H and Poudel, Rudra PK and Tsagkrasoulis, Dimosthenis and Caan, Matthan WA and Steves, Claire and Spector, Tim D and Montana, Giovanni},
  journal={NeuroImage},
  volume={163},
  pages={115--124},
  year={2017},
  publisher={Elsevier}
}

@article{peng2021accurate,
  title={Accurate brain age prediction with lightweight deep neural networks},
  author={Peng, Han and Gong, Weikang and Beckmann, Christian F and Vedaldi, Andrea and Smith, Stephen M},
  journal={Medical image analysis},
  volume={68},
  pages={101871},
  year={2021},
  publisher={Elsevier}
}

@article{bashyam2020mri,
  title={MRI signatures of brain age and disease over the lifespan based on a deep brain network and 14 468 individuals worldwide},
  author={Bashyam, Vishnu M and Erus, Guray and Doshi, Jimit and Habes, Mohamad and Nasrallah, Ilya M and Truelove-Hill, Monica and Srinivasan, Dhivya and Mamourian, Liz and Pomponio, Raymond and Fan, Yong and others},
  journal={Brain},
  volume={143},
  number={7},
  pages={2312--2324},
  year={2020},
  publisher={Oxford University Press}
}

@article{kickingereder2016radiomic,
  title={Radiomic profiling of glioblastoma: identifying an imaging predictor of patient survival with improved performance over established clinical and radiologic risk models},
  author={Kickingereder, Philipp and Burth, Sina and Wick, Antje and G{\"o}tz, Michael and Eidel, Oliver and Schlemmer, Heinz-Peter and Maier-Hein, Klaus H and Wick, Wolfgang and Bendszus, Martin and Radbruch, Alexander and others},
  journal={Radiology},
  volume={280},
  number={3},
  pages={880--889},
  year={2016},
  publisher={Radiological Society of North America}
}

@article{macyszyn2015imaging,
  title={Imaging patterns predict patient survival and molecular subtype in glioblastoma via machine learning techniques},
  author={Macyszyn, Luke and Akbari, Hamed and Pisapia, Jared M and Da, Xiao and Attiah, Mark and Pigrish, Vadim and Bi, Yingtao and Pal, Sharmistha and Davuluri, Ramana V and Roccograndi, Laura and others},
  journal={Neuro-oncology},
  volume={18},
  number={3},
  pages={417--425},
  year={2015},
  publisher={Society for Neuro-Oncology}
}

@inproceedings{weninger2018segmentation,
  title={Segmentation of brain tumors and patient survival prediction: Methods for the brats 2018 challenge},
  author={Weninger, Leon and Rippel, Oliver and Koppers, Simon and Merhof, Dorit},
  booktitle={International MICCAI brainlesion workshop},
  pages={3--12},
  year={2018},
  organization={Springer}
}

@inproceedings{caron2021emerging,
  title={Emerging properties in self-supervised vision transformers},
  author={Caron, Mathilde and Touvron, Hugo and Misra, Ishan and J{\'e}gou, Herv{\'e} and Mairal, Julien and Bojanowski, Piotr and Joulin, Armand},
  booktitle={Proceedings of the IEEE/CVF international conference on computer vision},
  pages={9650--9660},
  year={2021}
}

@article{oquab2023dinov2,
  title={Dinov2: Learning robust visual features without supervision},
  author={Oquab, Maxime and Darcet, Timoth{\'e}e and Moutakanni, Th{\'e}o and Vo, Huy and Szafraniec, Marc and Khalidov, Vasil and Fernandez, Pierre and Haziza, Daniel and Massa, Francisco and El-Nouby, Alaaeldin and others},
  journal={arXiv preprint arXiv:2304.07193},
  year={2023}
}

@article{simeoni2025dinov3,
  title={Dinov3},
  author={Sim{\'e}oni, Oriane and Vo, Huy V and Seitzer, Maximilian and Baldassarre, Federico and Oquab, Maxime and Jose, Cijo and Khalidov, Vasil and Szafraniec, Marc and Yi, Seungeun and Ramamonjisoa, Micha{\"e}l and others},
  journal={arXiv preprint arXiv:2508.10104},
  year={2025}
}

@inproceedings{chen2020simple,
  title={A simple framework for contrastive learning of visual representations},
  author={Chen, Ting and Kornblith, Simon and Norouzi, Mohammad and Hinton, Geoffrey},
  booktitle={International conference on machine learning},
  pages={1597--1607},
  year={2020},
  organization={PmLR}
}

@article{chen2020improved,
  title={Improved baselines with momentum contrastive learning},
  author={Chen, Xinlei and Fan, Haoqi and Girshick, Ross and He, Kaiming},
  journal={arXiv preprint arXiv:2003.04297},
  year={2020}
}

@article{grill2020bootstrap,
  title={Bootstrap your own latent-a new approach to self-supervised learning},
  author={Grill, Jean-Bastien and Strub, Florian and Altch{\'e}, Florent and Tallec, Corentin and Richemond, Pierre and Buchatskaya, Elena and Doersch, Carl and Avila Pires, Bernardo and Guo, Zhaohan and Gheshlaghi Azar, Mohammad and others},
  journal={Advances in neural information processing systems},
  volume={33},
  pages={21271--21284},
  year={2020}
}

@inproceedings{he2022masked,
  title={Masked autoencoders are scalable vision learners},
  author={He, Kaiming and Chen, Xinlei and Xie, Saining and Li, Yanghao and Doll{\'a}r, Piotr and Girshick, Ross},
  booktitle={Proceedings of the IEEE/CVF conference on computer vision and pattern recognition},
  pages={16000--16009},
  year={2022}
}

@inproceedings{xie2022simmim,
  title={Simmim: A simple framework for masked image modeling},
  author={Xie, Zhenda and Zhang, Zheng and Cao, Yue and Lin, Yutong and Bao, Jianmin and Yao, Zhuliang and Dai, Qi and Hu, Han},
  booktitle={Proceedings of the IEEE/CVF conference on computer vision and pattern recognition},
  pages={9653--9663},
  year={2022}
}

@article{bao2021beit,
  title={Beit: Bert pre-training of image transformers},
  author={Bao, Hangbo and Dong, Li and Piao, Songhao and Wei, Furu},
  journal={arXiv preprint arXiv:2106.08254},
  year={2021}
}

@inproceedings{kaczmarek2025building,
  title={Building a General SimCLR Self-Supervised Foundation Model Across Neurological Diseases to Advance 3D Brain MRI Diagnoses},
  author={Kaczmarek, Emily and Szeto, Justin and Nichyporuk, Brennan and Arbel, Tal},
  booktitle={Proceedings of the IEEE/CVF International Conference on Computer Vision},
  pages={1310--1319},
  year={2025}
}

@article{tak2026generalizable,
  title={A generalizable foundation model for analysis of human brain MRI},
  author={Tak, Divyanshu and Garomsa, Biniam A and Zapaishchykova, Anna and Chaunzwa, Tafadzwa L and Climent Pardo, Juan Carlos and Ye, Zezhong and Zielke, John and Ravipati, Yashwanth and Pai, Suraj and Vajapeyam, Sri and others},
  journal={Nature Neuroscience},
  pages={1--12},
  year={2026},
  publisher={Nature Publishing Group US New York}
}

@article{munk2024amaes,
  title={Amaes: Augmented masked autoencoder pretraining on public brain mri data for 3d-native segmentation},
  author={Munk, Asbj{\o}rn and Ambsdorf, Jakob and Llambias, Sebastian and Nielsen, Mads},
  journal={arXiv preprint arXiv:2408.00640},
  year={2024}
}

@article{robinet2025multimodal,
  title={Multimodal Masked Autoencoder Pre-training for 3D MRI-Based Brain Tumor Analysis with Missing Modalities},
  author={Robinet, Lucas and Berjaoui, Ahmad and Moyal, Elizabeth Cohen-Jonathan},
  journal={arXiv preprint arXiv:2505.00568},
  year={2025}
}

@article{mazher2025towards,
  title={Towards Generalisable Foundation Models for Brain MRI},
  author={Mazher, Moona and Parker, Geoff JM and Alexander, Daniel C},
  journal={arXiv preprint arXiv:2510.23415},
  year={2025}
}

@inproceedings{rui2025multi,
  title={Multi-modal vision pre-training for medical image analysis},
  author={Rui, Shaohao and Chen, Lingzhi and Tang, Zhenyu and Wang, Lilong and Liu, Mianxin and Zhang, Shaoting and Wang, Xiaosong},
  booktitle={Proceedings of the Computer Vision and Pattern Recognition Conference},
  pages={5164--5174},
  year={2025}
}

@article{yang2025genbrain,
  title={GenBrain: A Generative Foundation Model of Multimodal Brain Imaging},
  author={Yang, Chang and Feng, Jianfeng and Beckmann, Christian F and Smith, Stephen M and Gong, Weikang},
  journal={medRxiv},
  pages={2025--12},
  year={2025},
  publisher={Cold Spring Harbor Laboratory Press}
}

@article{guan2021domain,
  title={Domain adaptation for medical image analysis: a survey},
  author={Guan, Hao and Liu, Mingxia},
  journal={IEEE Transactions on Biomedical Engineering},
  volume={69},
  number={3},
  pages={1173--1185},
  year={2021},
  publisher={IEEE}
}

@article{yoon2024domain,
  title={Domain generalization for medical image analysis: A review},
  author={Yoon, Jee Seok and Oh, Kwanseok and Shin, Yooseung and Mazurowski, Maciej A and Suk, Heung-Il},
  journal={Proceedings of the IEEE},
  volume={112},
  number={10},
  pages={1583--1609},
  year={2024},
  publisher={IEEE}
}

@article{liu2023deep,
  title={Deep learning based brain tumor segmentation: a survey},
  author={Liu, Zhihua and Tong, Lei and Chen, Long and Jiang, Zheheng and Zhou, Feixiang and Zhang, Qianni and Zhang, Xiangrong and Jin, Yaochu and Zhou, Huiyu},
  journal={Complex \& intelligent systems},
  volume={9},
  number={1},
  pages={1001--1026},
  year={2023},
  publisher={Springer}
}

@article{allah2023edge,
  title={Edge U-Net: Brain tumor segmentation using MRI based on deep U-Net model with boundary information},
  author={Allah, Ahmed M Gab and Sarhan, Amany M and Elshennawy, Nada M},
  journal={Expert Systems with Applications},
  volume={213},
  pages={118833},
  year={2023},
  publisher={Elsevier}
}

@article{frisoni2010clinical,
  title={The clinical use of structural MRI in Alzheimer disease},
  author={Frisoni, Giovanni B and Fox, Nick C and Jack Jr, Clifford R and Scheltens, Philip and Thompson, Paul M},
  journal={Nature reviews neurology},
  volume={6},
  number={2},
  pages={67--77},
  year={2010},
  publisher={Nature Publishing Group UK London}
}

@article{feng2022deep,
  title={A deep learning MRI approach outperforms other biomarkers of prodromal Alzheimer’s disease},
  author={Feng, Xinyang and Provenzano, Frank A and Small, Scott A and Alzheimer’s Disease Neuroimaging Initiative},
  journal={Alzheimer's research \& therapy},
  volume={14},
  number={1},
  pages={45},
  year={2022},
  publisher={Springer}
}

@article{sarica2023explainability,
  title={Explainability of random survival forests in predicting conversion risk from mild cognitive impairment to Alzheimer’s disease},
  author={Sarica, Alessia and Aracri, Federica and Bianco, Maria Giovanna and Arcuri, Fulvia and Quattrone, Andrea and Quattrone, Aldo and Alzheimer’s Disease Neuroimaging Initiative},
  journal={Brain informatics},
  volume={10},
  number={1},
  pages={31},
  year={2023},
  publisher={Springer}
}

@article{zhang2022diagnosis,
  title={Diagnosis of Alzheimer's disease based on regional attention with sMRI gray matter slices},
  author={Zhang, Yanteng and Teng, Qizhi and Liu, Yuyang and Liu, Yan and He, Xiaohai},
  journal={Journal of neuroscience methods},
  volume={365},
  pages={109376},
  year={2022},
  publisher={Elsevier}
}

@article{liew2018large,
  title={A large, open source dataset of stroke anatomical brain images and manual lesion segmentations},
  author={Liew, Sook-Lei and Anglin, Julia M and Banks, Nick W and Sondag, Matt and Ito, Kaori L and Kim, Hosung and Chan, Jennifer and Ito, Joyce and Jung, Connie and Khoshab, Nima and others},
  journal={Scientific data},
  volume={5},
  number={1},
  pages={180011},
  year={2018},
  publisher={Nature Publishing Group}
}

@article{bommasani2021opportunities,
  title={On the opportunities and risks of foundation models},
  author={Bommasani, Rishi and Hudson, Drew A and Adeli, Ehsan and Altman, Russ and Arora, Simran and von Arx, Sydney and Bernstein, Michael S and Bohg, Jeannette and Bosselut, Antoine and Brunskill, Emma and others},
  journal={arXiv preprint arXiv:2108.07258},
  year={2021}
}

@inproceedings{kornblith2019similarity,
  title={Similarity of neural network representations revisited},
  author={Kornblith, Simon and Norouzi, Mohammad and Lee, Honglak and Hinton, Geoffrey},
  booktitle={International conference on machine learning},
  pages={3519--3529},
  year={2019},
  organization={PMlR}
}

@article{luckett2023predicting,
  title={Predicting survival in glioblastoma with multimodal neuroimaging and machine learning},
  author={Luckett, Patrick H and Olufawo, Michael and Lamichhane, Bidhan and Park, Ki Yun and Dierker, Donna and Verastegui, Gabriel Trevino and Yang, Peter and Kim, Albert H and Chheda, Milan G and Snyder, Abraham Z and others},
  journal={Journal of Neuro-oncology},
  volume={164},
  number={2},
  pages={309--320},
  year={2023},
  publisher={Springer}
}

@article{balakrishnan2019voxelmorph,
  title={Voxelmorph: a learning framework for deformable medical image registration},
  author={Balakrishnan, Guha and Zhao, Amy and Sabuncu, Mert R and Guttag, John and Dalca, Adrian V},
  journal={IEEE transactions on medical imaging},
  volume={38},
  number={8},
  pages={1788--1800},
  year={2019},
  publisher={IEEE}
}

@article{zhou2021ibot,
  title={ibot: Image bert pre-training with online tokenizer},
  author={Zhou, Jinghao and Wei, Chen and Wang, Huiyu and Shen, Wei and Xie, Cihang and Yuille, Alan and Kong, Tao},
  journal={arXiv preprint arXiv:2111.07832},
  year={2021}
}

@article{isensee2021nnu,
  title={nnU-Net: a self-configuring method for deep learning-based biomedical image segmentation},
  author={Isensee, Fabian and Jaeger, Paul F and Kohl, Simon AA and Petersen, Jens and Maier-Hein, Klaus H},
  journal={Nature methods},
  volume={18},
  number={2},
  pages={203--211},
  year={2021},
  publisher={Nature Publishing Group US New York}
}

@article{eidex2024deep,
  title={Deep learning in MRI-guided radiation therapy: A systematic review},
  author={Eidex, Zach and Ding, Yifu and Wang, Jing and Abouei, Elham and Qiu, Richard LJ and Liu, Tian and Wang, Tonghe and Yang, Xiaofeng},
  journal={Journal of applied clinical medical physics},
  volume={25},
  number={2},
  pages={e14155},
  year={2024},
  publisher={Wiley Online Library}
}

@article{li2025meddinov3,
  title={MedDINOv3: How to adapt vision foundation models for medical image segmentation?},
  author={Li, Yuheng and Wu, Yizhou and Lai, Yuxiang and Hu, Mingzhe and Yang, Xiaofeng},
  journal={arXiv preprint arXiv:2509.02379},
  year={2025}
}

@article{mazziotta2001probabilistic,
  title={A probabilistic atlas and reference system for the human brain: International Consortium for Brain Mapping (ICBM)},
  author={Mazziotta, John and Toga, Arthur and Evans, Alan and Fox, Peter and Lancaster, Jack and Zilles, Karl and Woods, Roger and Paus, Tomas and Simpson, Gregory and Pike, Bruce and others},
  journal={Philosophical Transactions of the Royal Society of London. Series B: Biological Sciences},
  volume={356},
  number={1412},
  pages={1293--1322},
  year={2001},
  publisher={The Royal Society}
}

@article{snoek2021amsterdam,
  title={The Amsterdam Open MRI Collection, a set of multimodal MRI datasets for individual difference analyses},
  author={Snoek, Lukas and van der Miesen, Maite M and Beemsterboer, Tinka and Van Der Leij, Andries and Eigenhuis, Annemarie and Steven Scholte, H},
  journal={Scientific data},
  volume={8},
  number={1},
  pages={85},
  year={2021},
  publisher={Nature Publishing Group UK London}
}

@article{wang2022longitudinal,
  title={A longitudinal neuroimaging dataset on language processing in children ages 5, 7, and 9 years old},
  author={Wang, Jin and Lytle, Marisa N and Weiss, Yael and Yamasaki, Brianna L and Booth, James R},
  journal={Scientific Data},
  volume={9},
  number={1},
  pages={4},
  year={2022},
  publisher={Nature Publishing Group UK London}
}

@article{di2014autism,
  title={The autism brain imaging data exchange: towards a large-scale evaluation of the intrinsic brain architecture in autism},
  author={Di Martino, Adriana and Yan, Chao-Gan and Li, Qingyang and Denio, Erin and Castellanos, Francisco X and Alaerts, Kaat and Anderson, Jeffrey S and Assaf, Michal and Bookheimer, Susan Y and Dapretto, Mirella and others},
  journal={Molecular psychiatry},
  volume={19},
  number={6},
  pages={659--667},
  year={2014},
  publisher={Nature Publishing Group}
}

@article{marek2011parkinson,
  title={The Parkinson progression marker initiative (PPMI)},
  author={Marek, Kenneth and Jennings, Danna and Lasch, Shirley and Siderowf, Andrew and Tanner, Caroline and Simuni, Tanya and Coffey, Chris and Kieburtz, Karl and Flagg, Emily and Chowdhury, Sohini and others},
  journal={Progress in neurobiology},
  volume={95},
  number={4},
  pages={629--635},
  year={2011},
  publisher={Elsevier}
}

@article{scarpace2016cancer,
  title={The cancer genome atlas glioblastoma multiforme collection (TCGA-GBM)},
  author={Scarpace, Lisa and Mikkelsen, Tom and Cha, Soonmee and Rao, Sujaya and Tekchandani, Sangeeta and Gutman, David and Saltz, Joel H and Erickson, Bradley J and Pedano, Nancy and Flanders, Adam E and others},
  journal={The Cancer Imaging Archive},
  year={2016},
  publisher={The Cancer Imaging Archive}
}

@misc{cptac2018gbm,
  title={The {Clinical Proteomic Tumor Analysis Consortium Glioblastoma Multiforme Collection (CPTAC-GBM)}},
  author={{National Cancer Institute Clinical Proteomic Tumor Analysis Consortium (CPTAC)}},
  year={2018},
  note={Version 16},
  howpublished={The Cancer Imaging Archive},
  doi={10.7937/K9/TCIA.2018.3RJE41Q1}
}

@misc{shusharina2021glis,
  title={Glioma Image Segmentation for Radiotherapy: {RT} targets, barriers to cancer spread, and organs at risk ({GLIS-RT})},
  author={Shusharina, Nadya and Bortfeld, Thomas},
  year={2021},
  howpublished={The Cancer Imaging Archive},
  doi={10.7937/TCIA.T905-ZQ20}
}

@misc{scarpace2019rembrandt,
  title={Data From {REMBRANDT}},
  author={Scarpace, Lisa and Flanders, Adam E and Jain, Rajan and Mikkelsen, Tom and Andrews, David W},
  year={2019},
  howpublished={The Cancer Imaging Archive},
  doi={10.7937/K9/TCIA.2015.588OZUZB}
}

@article{puchalski2018anatomic,
  title={An anatomic transcriptional atlas of human glioblastoma},
  author={Puchalski, Ralph B and Shah, Nameeta and Miller, Jeremy and Dalley, Rachel and Nomura, Steve R and Yoon, Jae-Guen and Smith, Kimberly A and Lankerovich, Michael and Bertagnolli, Darren and Bickley, Kris and others},
  journal={Science},
  volume={360},
  number={6389},
  pages={660--663},
  year={2018},
  publisher={American Association for the Advancement of Science}
}

@article{labella2024multi,
  title={A multi-institutional meningioma MRI dataset for automated multi-sequence image segmentation},
  author={LaBella, Dominic and Khanna, Omaditya and McBurney-Lin, Shan and Mclean, Ryan and Nedelec, Pierre and Rashid, Arif S and Tahon, Nourel Hoda and Altes, Talissa and Baid, Ujjwal and Bhalerao, Radhika and others},
  journal={Scientific data},
  volume={11},
  number={1},
  pages={496},
  year={2024},
  publisher={Nature Publishing Group UK London}
}

@article{adewole2023brain,
  title={The brain tumor segmentation (brats) challenge 2023: Glioma segmentation in sub-saharan africa patient population (brats-africa)},
  author={Adewole, Maruf and Rudie, Jeffrey D and Gbdamosi, Anu and Toyobo, Oluyemisi and Raymond, Confidence and Zhang, Dong and Omidiji, Olubukola and Akinola, Rachel and Suwaid, Mohammad Abba and Emegoakor, Adaobi and others},
  journal={ArXiv},
  pages={arXiv--2305},
  year={2023}
}

@article{juvekar2024remind,
  title={Remind: The brain resection multimodal imaging database},
  author={Juvekar, Parikshit and Dorent, Reuben and K{\"o}gl, Fryderyk and Torio, Erickson and Barr, Colton and Rigolo, Laura and Galvin, Colin and Jowkar, Nick and Kazi, Anees and Haouchine, Nazim and others},
  journal={Scientific data},
  volume={11},
  number={1},
  pages={494},
  year={2024},
  publisher={Nature Publishing Group UK London}
}

@article{zbontar2018fastmri,
  title={fastMRI: An open dataset and benchmarks for accelerated MRI},
  author={Zbontar, Jure and Knoll, Florian and Sriram, Anuroop and Murrell, Tullie and Huang, Zhengnan and Muckley, Matthew J and Defazio, Aaron and Stern, Ruben and Johnson, Patricia and Bruno, Mary and others},
  journal={arXiv preprint arXiv:1811.08839},
  year={2018}
}

@misc{kinahan2019acrin,
  title={Data from {ACRIN-DSC-MR-Brain}},
  author={Kinahan, Paul and Muzi, Mark and Bialecki, Brian and Herman, Brenda and Coombs, Laura},
  year={2019},
  howpublished={The Cancer Imaging Archive},
  doi={10.7937/tcia.2019.zr1pjf4i}
}

@misc{gerstner2012acrin,
  title={ACRIN 6684 assessment of tumor hypoxia in glioblastoma using 18F-fluoromisonidazole with PET and MRI.},
  author={Gerstner, Elizabeth Robins and Zhang, Zheng and Fink, James R and Sorensen, Gregory A and L'Heureux, Darryl and Heckel, Martha L and Dunning, Bernadine and Muzi, Mark and Mankoff, David A and Barboriak, Daniel},
  year={2012},
  publisher={American Society of Clinical Oncology}
}

@article{de20242024,
  title={The 2024 brain tumor segmentation (brats) challenge: Glioma segmentation on post-treatment mri},
  author={de Verdier, Maria Correia and Saluja, Rachit and Gagnon, Louis and LaBella, Dominic and Baid, Ujjwall and Tahon, Nourel Hoda and Foltyn-Dumitru, Martha and Zhang, Jikai and Alafif, Maram and Baig, Saif and others},
  journal={arXiv preprint arXiv:2405.18368},
  year={2024}
}

@misc{vassantachart2023meningioma,
  title={Segmentation and Classification of Grade {I} and {II} Meningiomas from Magnetic Resonance Imaging: An Open Annotated Dataset ({Meningioma-SEG-CLASS})},
  author={Vassantachart, April and Cao, Ying and Shen, Zhaoqi and Cheng, Kenneth and Gribble, Matthew and Ye, Jong C and Zada, Gabriel and Hurth, Kevin and Mathew, Anna and Guzman, Steven and Yang, Wensha},
  year={2023},
  howpublished={The Cancer Imaging Archive},
  doi={10.7937/0TKV-1A36}
}

@article{isensee2019automated,
  title={Automated brain extraction of multisequence MRI using artificial neural networks},
  author={Isensee, Fabian and Schell, Marianne and Pflueger, Irada and Brugnara, Gianluca and Bonekamp, David and Neuberger, Ulf and Wick, Antje and Schlemmer, Heinz-Peter and Heiland, Sabine and Wick, Wolfgang and others},
  journal={Human brain mapping},
  volume={40},
  number={17},
  pages={4952--4964},
  year={2019},
  publisher={Wiley Online Library}
}

@article{baid2021rsna,
  title={The rsna-asnr-miccai brats 2021 benchmark on brain tumor segmentation and radiogenomic classification},
  author={Baid, Ujjwal and Ghodasara, Satyam and Mohan, Suyash and Bilello, Michel and Calabrese, Evan and Colak, Errol and Farahani, Keyvan and Kalpathy-Cramer, Jayashree and Kitamura, Felipe C and Pati, Sarthak and others},
  journal={arXiv preprint arXiv:2107.02314},
  year={2021}
}

@article{moawad2024brain,
  title={The brain tumor segmentation-metastases (brats-mets) challenge 2023: Brain metastasis segmentation on pre-treatment mri},
  author={Moawad, Ahmed W and Janas, Anastasia and Baid, Ujjwal and Ramakrishnan, Divya and Saluja, Rachit and Ashraf, Nader and Maleki, Nazanin and Jekel, Leon and Yordanov, Nikolay and Fehringer, Pascal and others},
  journal={arxiv},
  pages={arXiv--2306},
  year={2024}
}

@article{labella2023asnr,
  title={The asnr-miccai brain tumor segmentation (brats) challenge 2023: Intracranial meningioma},
  author={LaBella, Dominic and Adewole, Maruf and Alonso-Basanta, Michelle and Altes, Talissa and Anwar, Syed Muhammad and Baid, Ujjwal and Bergquist, Timothy and Bhalerao, Radhika and Chen, Sully and Chung, Verena and others},
  journal={arXiv preprint arXiv:2305.07642},
  year={2023}
}

@misc{bratsgoat2024,
  title = {{BraTS-ISBI} 2024 -- Generalizability Across Tumors Challenge ({BraTS-GoAT})},
  howpublished = {Synapse},
  note = {Synapse ID: syn52939291},
  url = {https://www.synapse.org/Synapse:syn52939291},
  year = {2024}
}

@article{petersen2010alzheimer,
  title={Alzheimer's disease Neuroimaging Initiative (ADNI) clinical characterization},
  author={Petersen, Ronald Carl and Aisen, Paul S and Beckett, Laurel A and Donohue, Michael C and Gamst, Anthony Collins and Harvey, Danielle J and Jack Jr, Clifford R and Jagust, William J and Shaw, Leslie M and Toga, Arthur W and others},
  journal={Neurology},
  volume={74},
  number={3},
  pages={201--209},
  year={2010},
  publisher={Lippincott Williams \& Wilkins}
}

@article{marcus2007open,
  title={Open Access Series of Imaging Studies (OASIS): cross-sectional MRI data in young, middle aged, nondemented, and demented older adults},
  author={Marcus, Daniel S and Wang, Tracy H and Parker, Jamie and Csernansky, John G and Morris, John C and Buckner, Randy L},
  journal={Journal of cognitive neuroscience},
  volume={19},
  number={9},
  pages={1498--1507},
  year={2007},
  publisher={MIT Press}
}

@misc{ixi,
  title = {{IXI} -- Information eXtraction from Images},
  howpublished = {\url{https://brain-development.org/ixi-dataset/}},
  year = {2015}
}

@article{richardson2019mri,
  title={MRI data of 3--12 year old children and adults during viewing of a short animated film},
  author={Richardson, Hilary and Lisandrelli, Grace and Riobueno-Naylor, Alexa and Saxe, Rebecca},
  journal={Openneuro},
  year={2019}
}

@article{liew2022large,
  title={A large, curated, open-source stroke neuroimaging dataset to improve lesion segmentation algorithms},
  author={Liew, Sook-Lei and Lo, Bethany P and Donnelly, Miranda R and Zavaliangos-Petropulu, Artemis and Jeong, Jessica N and Barisano, Giuseppe and Hutton, Alexandre and Simon, Julia P and Juliano, Julia M and Suri, Anisha and others},
  journal={Scientific data},
  volume={9},
  number={1},
  pages={320},
  year={2022},
  publisher={Nature Publishing Group UK London}
}

@misc{ucsfpdgm,
  author = {Calabrese, Evan and Villanueva-Meyer, Javier and Rudie, Jeffrey and Rauschecker, Andreas and Baid, Ujjwal and Bakas, Spyridon and Cha, Soonmee and Mongan, John and Hess, Christopher},
  title = {The University of California San Francisco Preoperative Diffuse Glioma {MRI} ({UCSF-PDGM})},
  year = {2022},
  howpublished = {The Cancer Imaging Archive},
  doi = {10.7937/tcia.bdgf-8v37},
  note = {Version 5}
}

@article{kazerooni2024brain,
  title={The brain tumor segmentation (BraTS) challenge 2023: focus on pediatrics (CBTN-CONNECT-DIPGR-ASNR-MICCAI BraTS-PEDs)},
  author={Kazerooni, Anahita Fathi and Khalili, Nastaran and Liu, Xinyang and Haldar, Debanjan and Jiang, Zhifan and Anwar, Syed Muhammed and Albrecht, Jake and Adewole, Maruf and Anazodo, Udunna and Anderson, Hannah and others},
  journal={ArXiv},
  pages={arXiv--2305},
  year={2024}
}

@misc{upenn-gbm,
  author = {Bakas, Spyridon and Sako, Chiharu and Akbari, Hamed and Bilello, Michel and Sotiras, Aristeidis and Shukla, Gaurav and Rudie, Jeffrey D. and Flores Santamaria, Nadia and Fathi Kazerooni, Anahita and Pati, Sarthak and Rathore, Saima and Mamourian, Elizabeth and Ha, Sung Min and Parker, William and Doshi, Jimit and Baid, Ujjwal and Bergman, Mark and Binder, Zev A. and Verma, Ragini and Davatzikos, Christos},
  title = {Multi-parametric magnetic resonance imaging ({mpMRI}) scans for de novo Glioblastoma ({GBM}) patients from the University of Pennsylvania Health System ({UPENN-GBM})},
  year = {2021},
  howpublished = {The Cancer Imaging Archive},
  doi = {10.7937/TCIA.709X-DN49},
  note = {Version 2}
}

@article{koirala2019alterations,
  title={Alterations in white matter network and microstructural integrity differentiate Parkinson’s disease patients and healthy subjects},
  author={Koirala, Nabin and Anwar, Abdul Rauf and Ciolac, Dumitru and Glaser, Martin and Pintea, Bogdan and Deuschl, G{\"u}nther and Muthuraman, Muthuraman and Groppa, Sergiu},
  journal={Frontiers in aging neuroscience},
  volume={11},
  pages={191},
  year={2019},
  publisher={Frontiers Media SA}
}

@inproceedings{wei2026fmri,
  title={fmri-lm: Towards a universal foundation model for language-aligned fmri understanding},
  author={Wei, Yuxiang and Zhang, Yanteng and Xiao, Xi and Qian, Chengxuan and Wang, Tianyang and Calhoun, Vince D},
  booktitle={Proceedings of the IEEE/CVF Conference on Computer Vision and Pattern Recognition},
  pages={6931--6940},
  year={2026}
}

@article{xiao2026brainomni,
  title={Brainomni: A brain foundation model for unified eeg and meg signals},
  author={Xiao, Qinfan and Cui, Ziyun and Zhang, Chi and Chen, Siqi and Wu, Wen and Thwaites, Andrew and Woolgar, Alexandra and Zhou, Bowen and Zhang, Chao},
  journal={Advances in Neural Information Processing Systems},
  volume={38},
  pages={41179--41212},
  year={2026}
}

@article{monachino2025self,
  title={Self-DANA: A Resource-Efficient Channel-Adaptive Self-Supervised Approach for ECG Foundation Models},
  author={Monachino, Giuliana and La Porta, Nicol{\`o} and Zanchi, Beatrice and Fiorillo, Luigi and Rossi, Alvise Dei and Farina, Georgiy and Faraci, Francesca Dalia},
  journal={arXiv preprint arXiv:2507.14151},
  year={2025}
}

\end{document}